\pdfoutput=1
\documentclass{article}

\usepackage{etoolbox}

\PassOptionsToPackage{numbers}{natbib}
\usepackage[final]{neurips_2024}

\usepackage[utf8]{inputenc} %
\usepackage[T1]{fontenc}    %
\usepackage{hyperref}       %
\usepackage{url}            %
\usepackage{booktabs}       %
\usepackage{amsfonts}       %
\usepackage{nicefrac}       %
\usepackage{microtype}      %
\usepackage{xcolor}         %

\usepackage{wrapfig}
\usepackage{amsmath}
\usepackage{amssymb}
\usepackage{amsthm}
\usepackage{booktabs}

\usepackage{enumitem}
\usepackage{makecell}
\usepackage{diagbox}
\usepackage{bm}
\usepackage{dsfont}
\usepackage{subfig}
\usepackage{multirow}
\usepackage{varwidth}
\usepackage{pifont}
\usepackage{makecell}
\usepackage{graphicx} 
\usepackage{comment}
\usepackage{color}
\usepackage[colaction]{multicol}
\usepackage{colortbl}
\usepackage{algorithm}%
\usepackage{xspace}
\usepackage{rotating}

\usepackage{adjustbox}
\usepackage{threeparttable}
\usepackage{listings}

\let\classAND\AND
\let\AND\relax
\let\classOR\OR
\let\OR\relax
\usepackage{algorithmic}

\let\AND\classAND

\let\OR\classOR
\AtBeginEnvironment{algorithmic}{\let\AND\algoAND}
\AtBeginEnvironment{algorithmic}{\let\OR\algoOR}

\floatname{algorithm}{Algorithm}

\definecolor{gc_pink}{HTML}{FF6F79}
\definecolor{gc_blue}{HTML}{6FB0FF}
\definecolor{gc_gray}{HTML}{D9D9D9}
\definecolor{gc_dark_pink}{HTML}{99262E}
\definecolor{gc_dark_blue}{HTML}{265A99}
\definecolor{gc_dark_gray}{HTML}{999999}
\definecolor{comment_color}{HTML}{1B8F44}
\definecolor{comment_color_2}{RGB}{64,128,128}

\newcommand{\method}{MInference}
\newcommand{\methodall}{MInference 1.0}
\newcommand{\LineComment}[1]{\vspace*{0.5em}\small\textcolor{comment_color_2}{\textit{\# #1}}}

\definecolor{codegreen}{rgb}{0,0.6,0}
\definecolor{codegray}{rgb}{0.5,0.5,0.5}
\definecolor{codepurple}{rgb}{0.58,0,0.82}
\definecolor{backcolour}{rgb}{0.95,0.95,0.92}

\lstdefinestyle{mystyle}{
    backgroundcolor=\color{backcolour},   
    commentstyle=\color{codegreen},
    keywordstyle=\color{magenta},
    numberstyle=\tiny\color{codegray},
    stringstyle=\color{codepurple},
    basicstyle=\ttfamily\footnotesize,
    breakatwhitespace=false,         
    breaklines=true,                 
    captionpos=false,                    
    keepspaces=true,                 
    numbersep=5pt,                  
    showspaces=false,                
    showstringspaces=false,
    showtabs=false,                  
    tabsize=2
}

\usepackage{caption}
\usepackage{minitoc}

\newcommand{\wh}{\widehat}

\renewcommand{\hat}{\wh}

\DeclareMathOperator*{\AND}{\mathrm{AND}}
\DeclareMathOperator*{\OR}{\mathrm{OR}}

\makeatletter
\def\@fnsymbol#1{\ensuremath{\ifcase#1\or \dagger \or  \ddagger\or
   \mathsection\or  \text{*}\or \mathparagraph \or  \| \or **\or \dagger\dagger
   \or \ddagger\ddagger \else\@ctrerr\fi}}
\makeatother
\renewcommand{\thefootnote}{\fnsymbol{footnote}}

\title{\methodall{}: Accelerating Pre-filling for Long-Context LLMs via Dynamic Sparse Attention}

\author{
Huiqiang Jiang\footnotemark[1], Yucheng Li$^{\lozenge}$\footnotemark[1], Chengruidong Zhang\footnotemark[1], Qianhui Wu, Xufang Luo, \\
\bf Surin Ahn, Zhenhua Han, Amir H. Abdi, Dongsheng Li, Chin-Yew Lin, Yuqing Yang, Lili Qiu\\
Microsoft Corporation, $^{\lozenge}$University of Surrey\\
\texttt{\{hjiang,chengzhang,yuqyang\}@microsoft.com,yucheng.li@surrey.ac.uk}\\
}

\begin{document}

\maketitle
\footnotetext[1]{Equal contribution. $^{\lozenge}$Work during internship at Microsoft.}

\renewcommand{\thefootnote}{\arabic{footnote}}  %

\begin{abstract}
The computational challenges of Large Language Model (LLM) inference remain a significant barrier to their widespread deployment, especially as prompt lengths continue to increase.
Due to the quadratic complexity of the attention computation,
it takes 30 minutes for an 8B LLM to process a prompt of 1M tokens (i.e., the pre-filling stage) on a single A100 GPU.
Existing methods for speeding up pre-filling often fail to maintain acceptable accuracy or efficiency when applied to long-context LLMs. 
To address this gap, we introduce $\textbf{\method{}}$ (\textit{Million-tokens Inference}), a sparse calculation method designed to accelerate pre-filling of long-sequence processing. 
Specifically, we identify three unique patterns in long-context attention matrices---the \textit{A-shape}, \textit{Vertical-Slash}, and \textit{Block-Sparse}---that can be leveraged for efficient sparse computation on GPUs. We determine the optimal pattern for each attention head offline and dynamically build sparse indices based on the assigned pattern during inference. With the pattern and sparse indices, we perform efficient sparse attention calculations via our optimized GPU kernels to significantly reduce the latency in the pre-filling stage of long-context LLMs.
Our proposed technique can be directly applied to existing LLMs without any modifications to the pre-training setup or additional fine-tuning. By evaluating on a wide range of downstream tasks, including InfiniteBench, RULER, PG-19, and Needle In A Haystack, and models including LLaMA-3-1M, GLM-4-1M, Yi-200K, Phi-3-128K, and Qwen2-128K, we demonstrate that \method{} effectively reduces inference latency by up to $10\times$ for pre-filling on an A100, while maintaining accuracy.
Our code is available at \url{https://aka.ms/MInference}.

\end{abstract}

\begin{figure*}[htb]
  \vspace{-14pt}
  \centering
  \subfloat[Needle In A Haystack]{
    \label{sfig:needle_ours_result}
    \includegraphics[width=0.49\columnwidth]{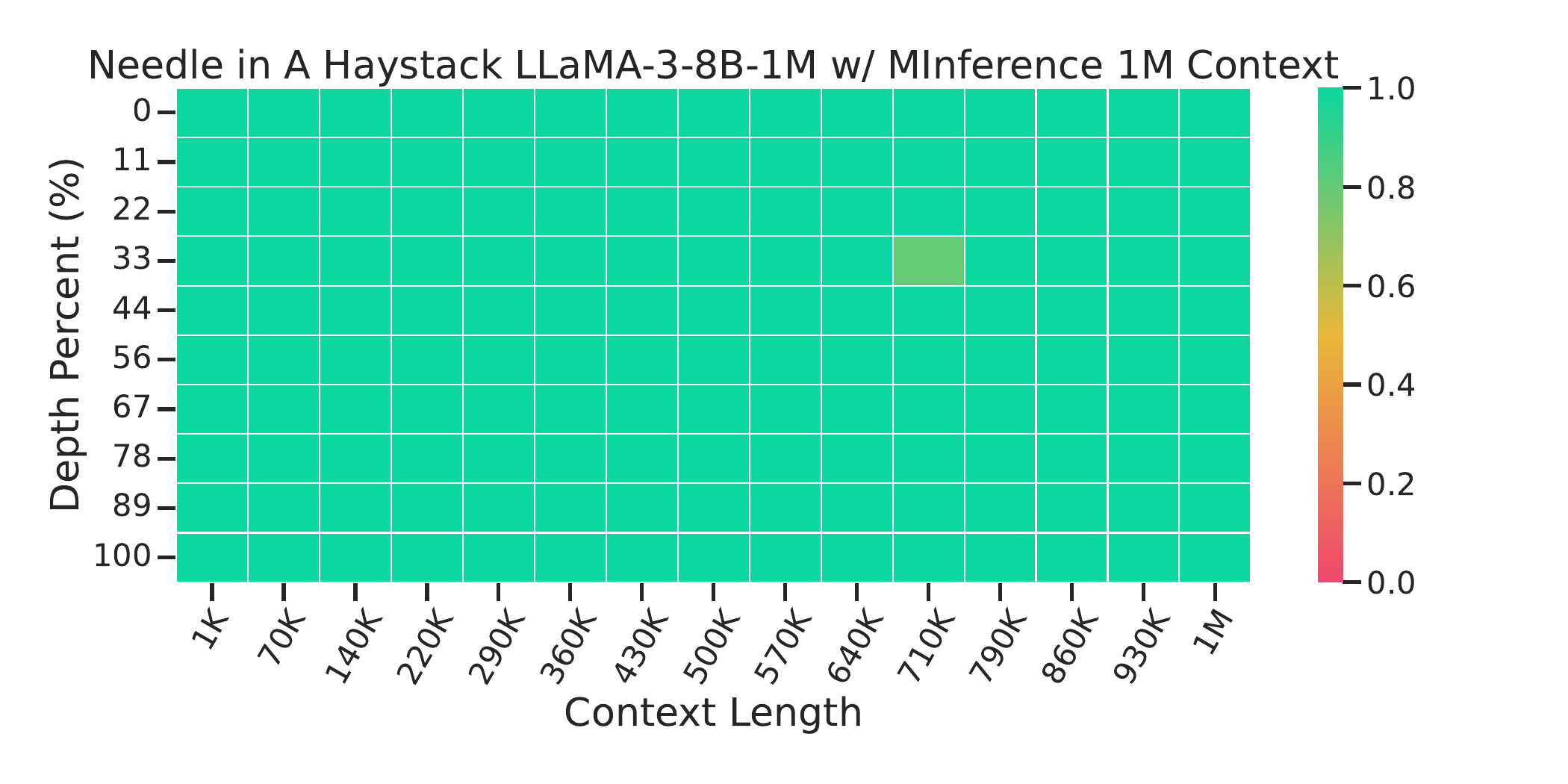}}
  \subfloat[Latency Speedup]{
    \label{sfig:speedup}
    \includegraphics[width=0.49\columnwidth]{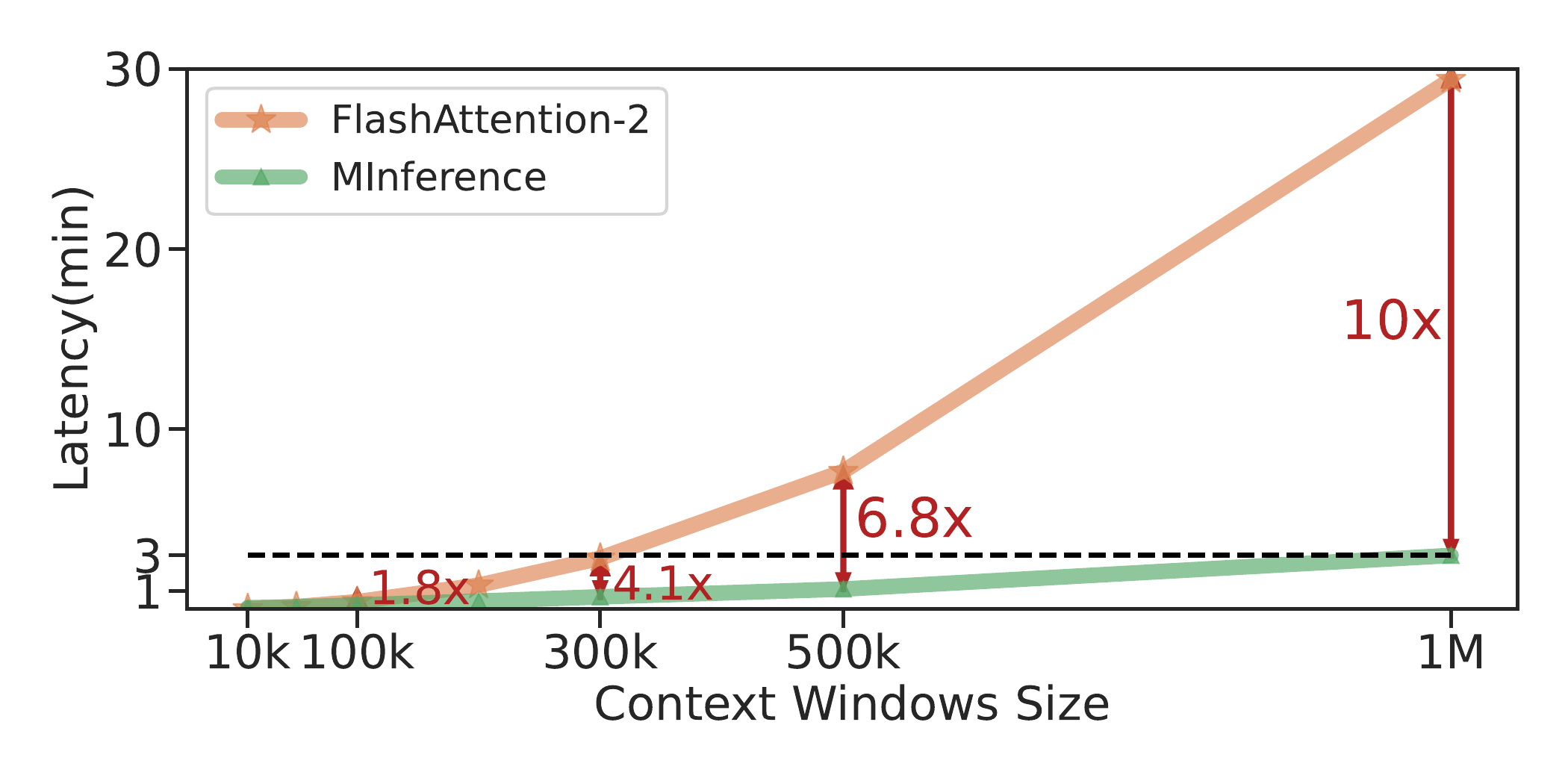}}
  \caption{Attention weights, especially in long-context LLMs, exhibit up to 96.8\% sparsity in contexts of 128K. We propose \textbf{\method{}}, leveraging dynamic sparse attention to accelerate the pre-filling stage of long-context LLM inference. It achieves up to 10x speedup for 1M contexts on a single A100, as shown in (b), and matches or surpasses baselines, as demonstrated by Needle In A Haystack~\cite{kamradt2023needle} in (a) on LLaMA-3-8B-1M~\cite{gradient2024}.
  }
  \label{fig:performance_gain}
\end{figure*}

\section{Introduction}

Large language models (LLMs) have entered the era of long-context processing, with some of them supporting context windows ranging from 128K to 10M tokens~\cite{gradient2024,reid2024gemini,liu2023world,young2024yi,Abdin2024Phi3TR,deepseekv2}. These extended context windows enable LLMs to unlock a multitude of complex real-world applications, such as repository-level code understanding~\cite{bairi2023codeplan,Jimenez2023SWEbenchCL,Park2023GenerativeAI}, long-document question-answering~\cite{caciularu2023peek,li2024long}, self-play reasoning~\cite{openai2024o1}, extreme-label in-context learning~\cite{li2024long}, and long-horizon agent tasks~\cite{weng2023agent}.

\begin{figure*}[htb]
  \centering
  \subfloat[Attention incurs heavy cost.]{
    \label{sfig:latency}
    \includegraphics[height=0.26\columnwidth]{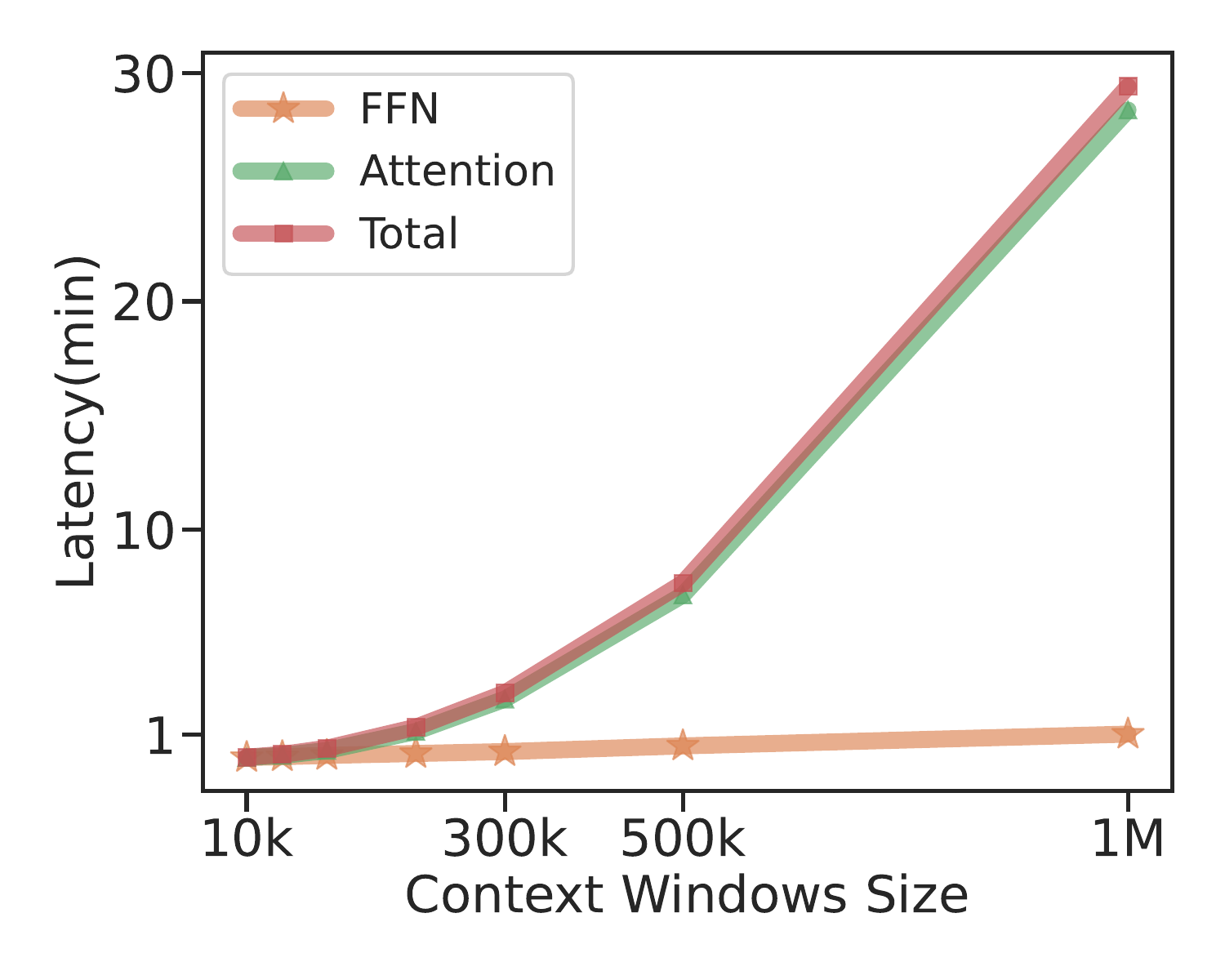}}
  \subfloat[Attention is sparse.]{
    \label{sfig:attention_is_sparse}
    \includegraphics[height=0.26\columnwidth]{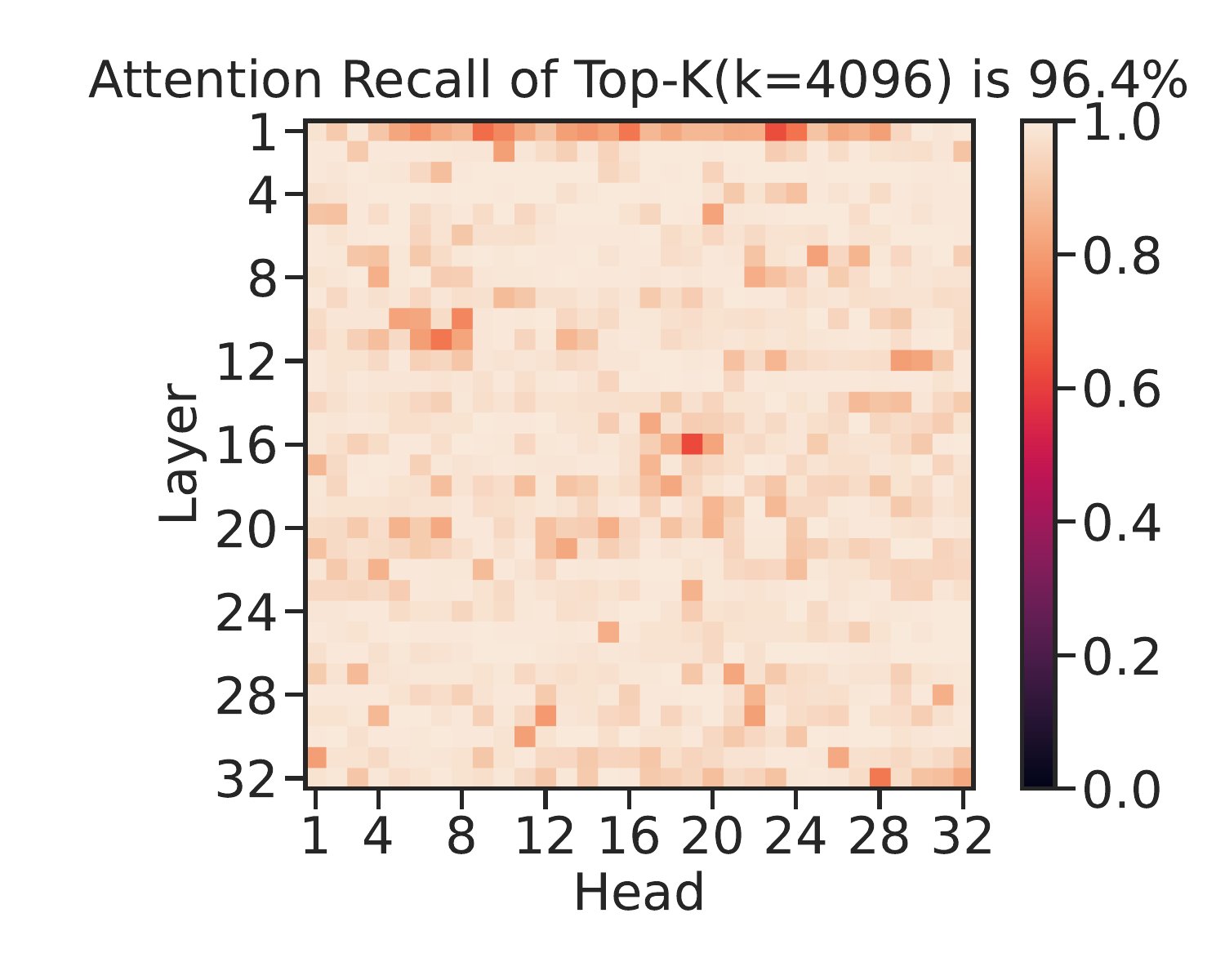}}
  \subfloat[Sparsity of attention is dynamic.]{
    \label{sfig:attention_sparisity_is_dynamic}
    \includegraphics[height=0.26\columnwidth]{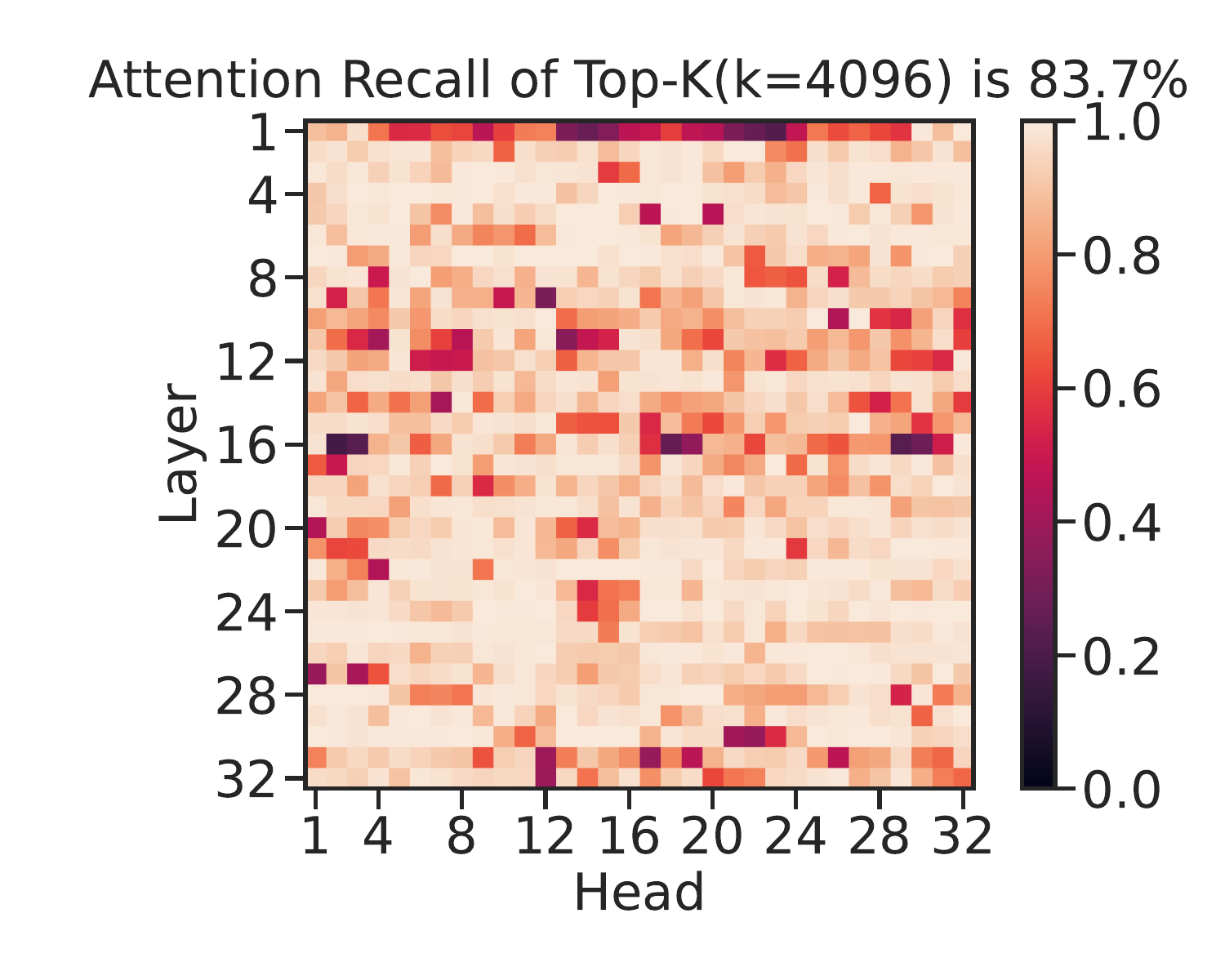}}
  \caption{(a) Latency breakdown of the pre-filling stage. (b) How much attention scores can top-k (k=4096) columns cover in a 128k context. (c) Less attention scores are retrieved when reusing the top-k indices from another examples, indicating its dynamic nature. Visualizations are based on LLaMa-3-8B with a single A100.}
  \label{fig:motivations_sparsity}
  \vspace{-10pt}
\end{figure*}

However, due to the quadratic complexity of attention, it can take several minutes for the model to process the input prompt (i.e., the pre-filling stage) and then start to produce the first token, which leads to unacceptable Time To First Token experience,
thus greatly hinders the wide application of long-context LLMs.
As shown in Fig.~\ref{sfig:latency}, when serving LLaMA-3-8B on a single A100 machine, the model would keep users waiting for 6 minutes to finish the pre-filling stage given a prompt of 300K tokens, and this number increases to 30 minutes for a prompt of 1M tokens. The overhead of self-attention computation exceeds 90\% of the total pre-filling latency, which makes it the major bottleneck in long-context processing of LLMs.
Previous research has shown that the attention matrices are highly sparse \cite{liu2021transformer,deng2024attention}, which has led to the development of fixed sparse attention methods such as Longformer~\cite{beltagy2020longformer} and BigBird~\cite{zaheer2020big}. However, prior studies have also noted that attention distributions vary significantly across different inputs \cite{likhosherstov2021expressive,liu2021transformer}. This dynamic nature prevents prior sparse methods from being used directly on long-context LLMs without expensive training or fine-tuning.
But if the dynamic sparse attention patterns could be efficiently predicted online, the pre-filling latency of long-context LLMs could be significantly reduced by calculating only the most important part of the attention weights.

Building upon this idea, we present \textbf{\method{}}, a technique that reduces 95\% of FLOPs in the attention computation to significantly accelerate the pre-filling stage of long-context LLM inference via dynamic sparse attention. Unlike existing dynamic sparse attention methods that introduce large computational overhead to estimate attention patterns with low-rank hidden dimensions~\cite{liu2021transformer,ribar2023sparq}, 
our method is designed specifically for long-context scenarios with minimal overhead in estimation.
Specifically, we conduct extensive analysis and identify three general patterns of sparse attention in long-context LLMs: \textit{A-shape} pattern, \textit{Vertical-Slash} pattern, and \textit{Block-Sparse} pattern. Based on these findings, we introduce a kernel-aware search method to assign the optimal attention pattern for each head. Importantly, instead of fixed attention masks in prior studies, we perform an efficient online approximation to build a dynamic sparse mask for each head according to their assigned pattern and particular inputs.
For example, to build a dynamic sparse mask for a specific prompt on one \textit{Vertical-Slash} head, we use a partial of attention weight consisting of the last $\text{last\_q}$ query 
and key vectors (i.e. $\bm{Q}_{[-\text{last\_q}:]}$ and $\bm{K}$) to estimate the most important indices of the vertical and slash lines globally on the attention matrix.
For \textit{Block-Sparse} heads, we perform mean pooling on both query and key vectors in blocks of 64 and calculate the block-level attention weights to determine the most important blocks and thereby obtain a block-sparse dynamic mask. 
After obtaining the dynamic sparse mask, three optimized GPU kernels are used, which we developed for the above three sparse patterns. These kernels are based on the dynamic sparse compilers PIT~\cite{zheng2023pit}, Triton~\cite{tillet2019triton} and FlashAttention~\cite{dao2023flashattention}, which enable extremely efficient computation of dynamic sparse attention.

Extensive experiments are conducted on various Long-context LLMs, including LLaMA-3-8B-1M~\cite{gradient2024}, GLM-4-9B-1M~\cite{glm2024chatglm}, and Yi-9B-200K~\cite{young2024yi}, across benchmarks with context lengths over 1M tokens, such as InfiniteBench~\cite{zhang2024inftybench}, RULER~\cite{hsieh2024ruler}, Needle In A Haystack~\cite{kamradt2023needle}, and PG-19~\cite{rae2019compressive}. Needle In A Haystack was also tested on Phi-3-Mini-128K~\cite{Abdin2024Phi3TR} and Qwen-2-7B-128K~\cite{qwen}. Results show that \method{} speeds up the pre-filling stage by up to $10\times$ for 1M contexts with LLaMA-3-8B on a single A100, reducing latency from 30 minutes to 3 minutes per prompt, while maintaining or improving accuracy.

\section{Attention Heads: Dynamic, Sparse, and Characteristic}
\label{sec:motivation}

\subsection{Attention is Dynamically Sparse}

The sparsity of attention weights in pre-trained LLMs, especially in long-context scenarios, has been well-documented~\cite{liu2021transformer,ribar2023sparq,pmlr-v202-liu23am,xiao2023streamingllm}. As shown in Fig.~\ref{sfig:attention_is_sparse}, for an attention matrix of size $128k \times 128k$, retaining only the top 4k columns recalls 96.8\% of the total attention. In other words, each token is attending to a limit number of tokens despite the long sequence it is processing.

On the other hand, although the sparse nature of attention matrices is shared across different inputs, the exact distributions of sparse pattern are highly dynamic. 
That is to say, a token at a given position only attends to a subset of the sequence in self-attention, and the exact tokens it attends to are highly context-dependent and vary significantly across different prompts.
This dynamism has been mathematically demonstrated in prior studies~\cite{likhosherstov2021expressive,Likhosherstov_Choromanski_Weller_2023}. As depicted in Fig.~\ref{sfig:attention_sparisity_is_dynamic},
if we take the top 4k columns found in Fig.~\ref{sfig:attention_is_sparse} and apply it on another prompt of 128k, 
the recall of attention would drop largely to 83.7\%.

\subsection{Attention Sparsity Exhibits Patterns}
\label{subsec:attention_pattern}

\begin{table}[ht]
    \centering
    \vspace{-10pt}
    \caption{Comparison of different sparse patterns. 
    }
     \resizebox{0.95\columnwidth}{!}{
    \begin{tabular}{l|ccc|c}
    \toprule
         Patterns & A-shape & Vertical-Slash & Block-Sparse & Top-K  \\
         \midrule
        Spatial Distribution & Static structured & Dynamic structured & Dynamic structured & Dynamic fine-grained  \\
        Latency on GPU & Low & Medium & Low & High\\
        Time to build the index & Zero & Small & Small & High \\
        \bottomrule
    \end{tabular}
    }
    \label{tab:pattern}
\end{table}

\begin{figure*}[!b]
  \centering
  \vspace{-10pt}
  \subfloat[Attention patterns]{
    \label{sfig:attention_pattern}
    \includegraphics[height=0.3\columnwidth]{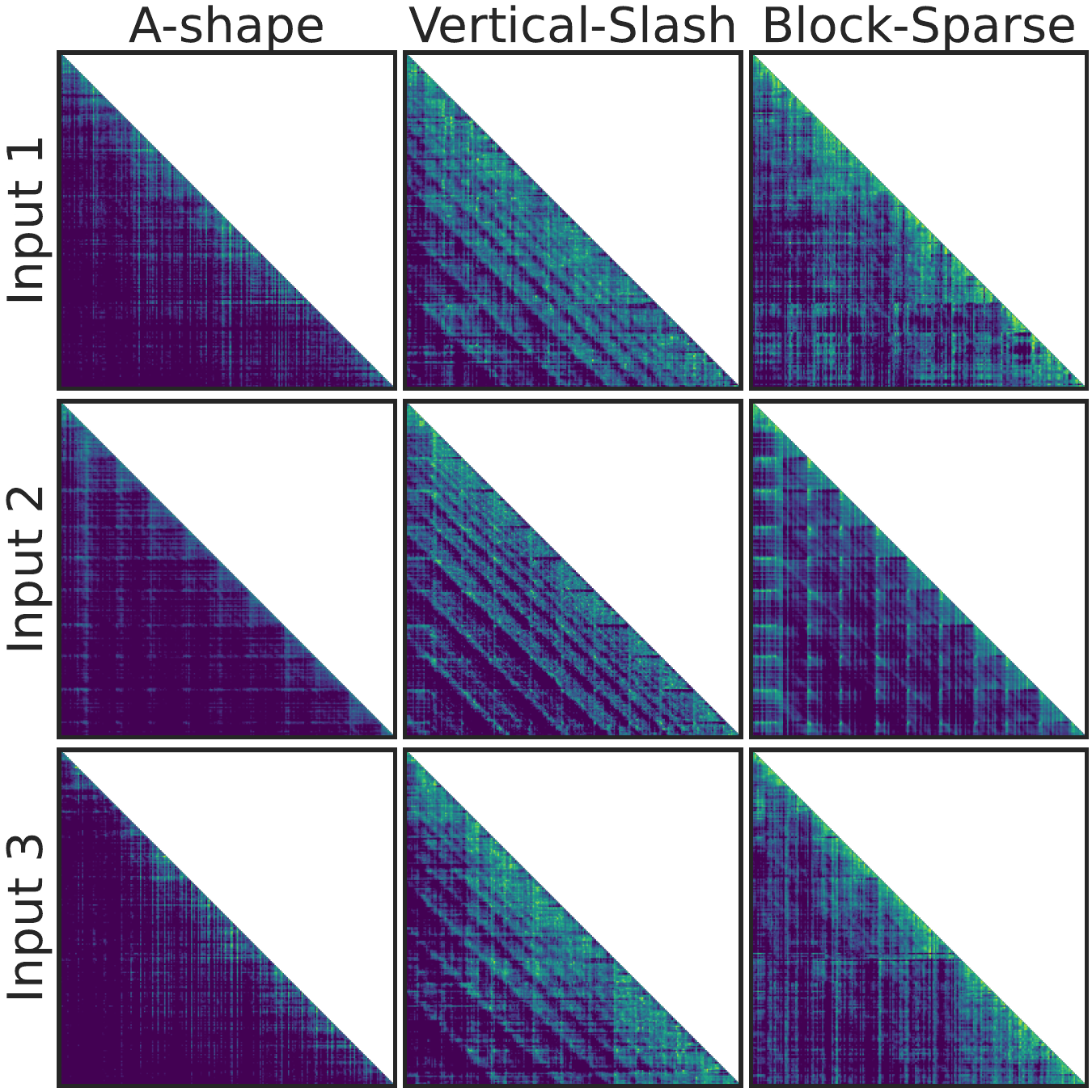}}
  \subfloat[Attention is spatial clustering]{
    \label{sfig:nearest_no_zero}
    \includegraphics[height=0.28\columnwidth]{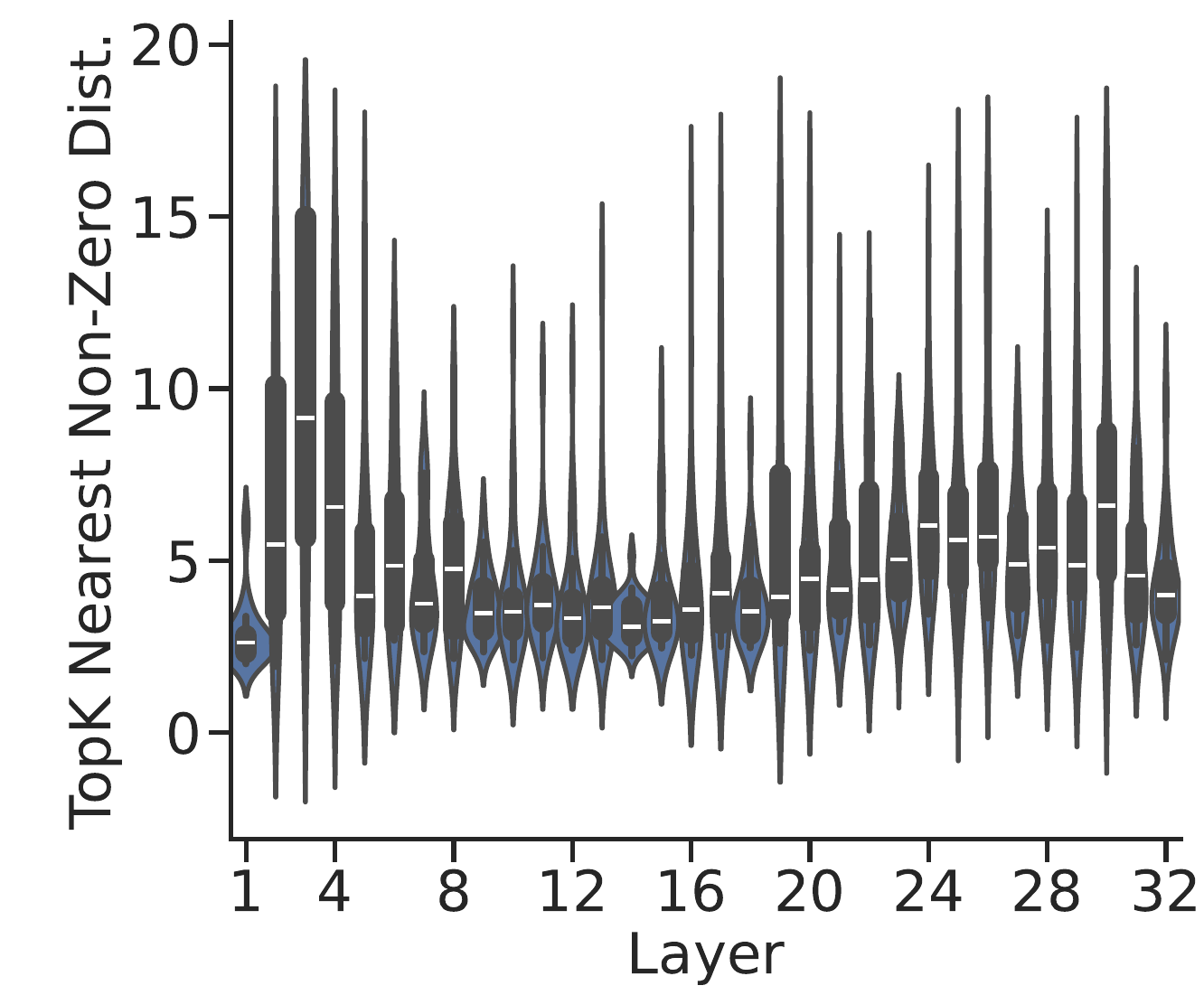}}
  \subfloat[Attention pattern recall]{
    \label{sfig:pattern_recall}
    \includegraphics[height=0.3\columnwidth]{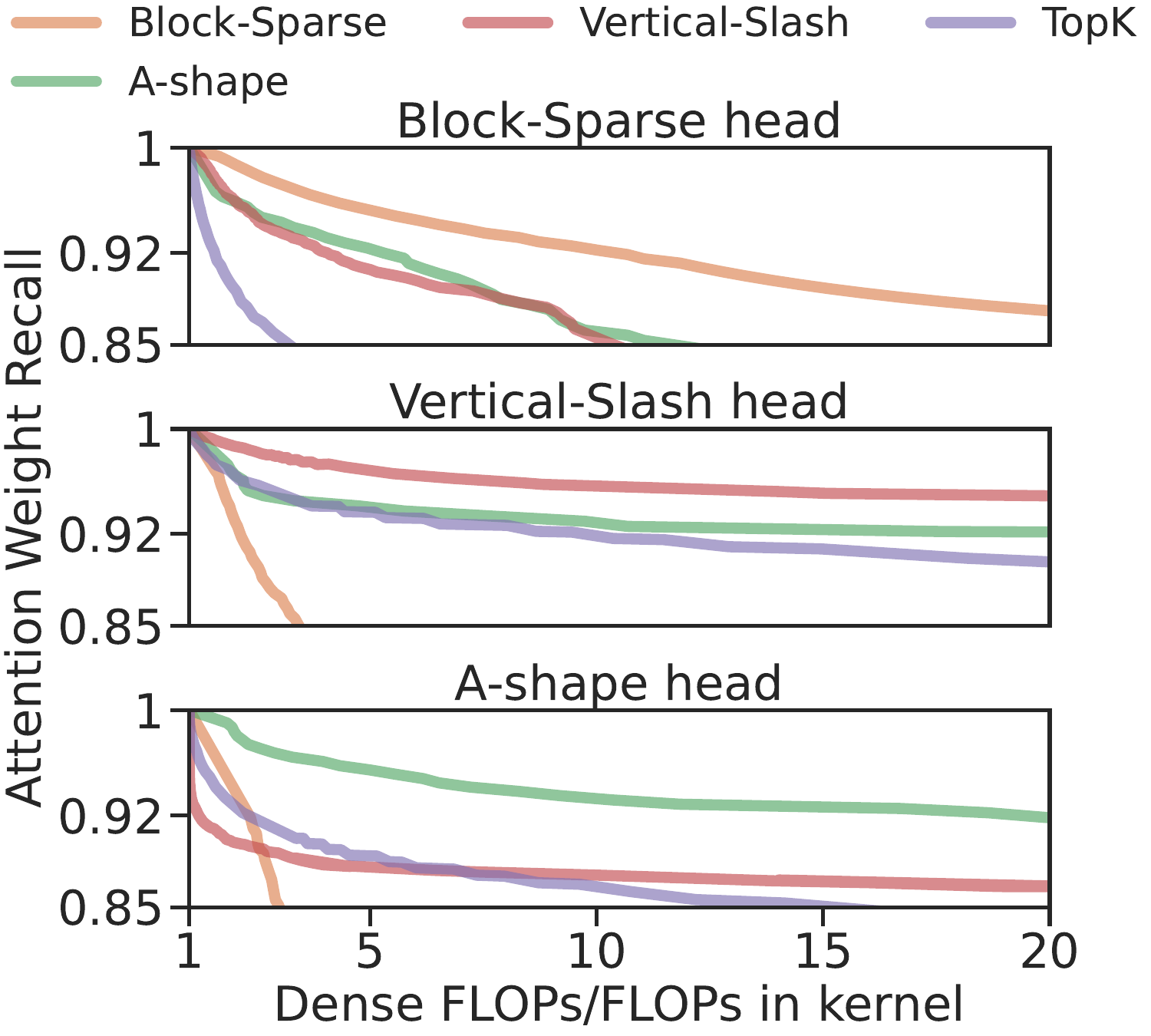}}
  \caption{(a) Visualization of attention weights from different attention heads. For different prompts and tasks, the pattern of the same head is relatively consistent, but the sparse indices are dynamically changing.(b) Distance of the top-10 nearest non-zero element in the attention matrix. (c) Attention recall distribution using our identified patterns, where FLOPs in the kernel refer to the real FLOPs required for sparse attention computing using on GPUs. Here, a 1x64 block size is used for the \textit{Vertical-Slash} pattern, and a 64x64 block size is used for others on GPUs. All visualization are based on LLaMA-3-8B-Instruct-262K~\cite{gradient2024}.
  }
  \label{fig:motivations_pattern}
\end{figure*}

\begin{figure*}[htb]
    \centering
    \resizebox{\columnwidth}{!}{
    \includegraphics[width=1\linewidth]{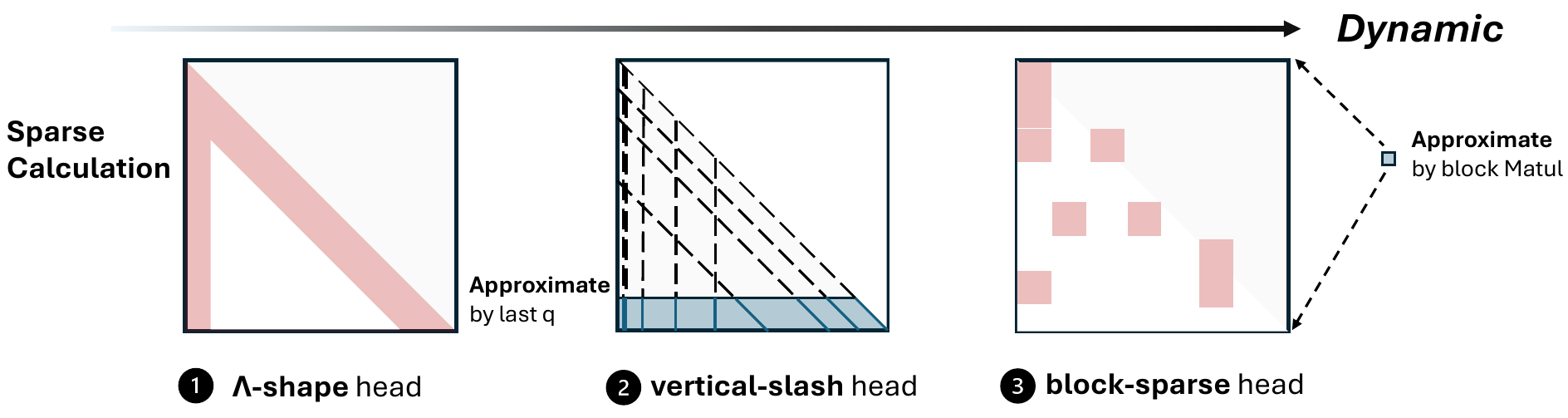}
    }
    \caption{The three sparse methods in \method{}.}
    \label{fig:framework}
\end{figure*}

Although the sparsity distribution of attention matrix is dynamic, previous works~\cite{xiao2023streamingllm,han2023lminfinite} have shown that they exhibit certain patterns in the two-dimensional space such as spatial clustering.
Through our analysis of long-context prompts of various lengths and tasks, we have categorized such attention sparse patterns into the \textit{A-shape}, \textit{Vertical-Slash} (VS), and \textit{Block-Sparse} patterns, as shown in Fig.~\ref{sfig:attention_pattern} and Fig.~\ref{fig:framework}.
Table~\ref{tab:pattern} details the characteristics and differences between these three patterns.

\textbf{\textit{A-shape} pattern} The attention weights of these types of heads are concentrated on initial tokens and local windows~\cite{xiao2023streamingllm,han2023lminfinite}, exhibiting relatively higher stability. 

\textbf{\textit{Vertical-Slash} (VS) pattern} The attention weights %
are concentrated on specific tokens (vertical lines)~\cite{mohtashami2023landmark} and tokens at fixed intervals (slash lines). The positions of vertical and slash lines in this pattern dynamically change with the context content and exhibit a certain sparsity, making them difficult to be encompassed by local windows and \textit{A-shape} patterns.

\textbf{\textit{Block-Sparse} pattern} This %
sparsity pattern is the most dynamic, exhibiting a more dispersed distribution. %
Despite its dynamism,
the attention weights maintain some characteristics of spatial clustering, which we identify as the block-sparse pattern. We analyzed the distances between non-zero attention weights and their top-k nearest non-zero neighbors within a 128k prompt as shown in Fig.~\ref{sfig:nearest_no_zero}. The results indicate that across layers and heads, the distances between nearest non-zero values are generally concentrated around 5, suggesting a strong spatial clustering of the attention weights.

The point of these three patterns is that we can leverage them to perform highly efficient sparse computing for the attention matrix in long-context LLMs. In Fig.~\ref{sfig:pattern_recall}, we test how efficient is our indentified patterns retrieving attention scores with limit computing cost on GPU (FLOPs). First, attention heads are labeled with one of the sparse pattern (detail see \S\ref{3.2}). Then we demonstrate our patterns are significantly more efficient compared to other sparse methods \cite{ribar2023sparq,xiao2023streamingllm,pagliardini2024fast}. Specifically, with the same amount of FLOPs, our patterns achieve a notable higher recall on attention scores, which can potentially lead to better accuracy.
For example, previous Top-K methods~\cite{ribar2023sparq,xiao2023streamingllm,pagliardini2024fast} struggle with the \textit{Block-Sparse} pattern as they focus on specific tokens globally, while our pattern retrieves attention scores more efficiently and accurately. We example how we use these patterns on long-context LLMs and how we implement optimized GPU kernels for these patterns in \S\ref{sec:sparseq}.

\section{\methodall{}}
\label{sec:sparseq}

Following the analysis in \S\ref{sec:motivation}, we propose \textbf{\method{}} to accelerate the pre-filling stage of long-context LLMs, consisting of three steps:
1) Offline attention pattern identification for each head; 2) Dynamic build of sparse indices w.r.t. the pattern; 3) Sparse attention calculation with optimized GPU kernels.

\subsection{Problem Formulation}

When accelerating the pre-filling stage of long-context LLMs with sparse attention computing, the attention matrix can be formulated as follows:
\begin{equation}
    \bm{A(M)} = \text{Softmax}(\frac{1}{\sqrt{d}}\bm{Q}\bm{K}^\top - c(1-\bm{M})),
\end{equation}
where $M_{i,j} \in \{0,1\}$ represents the dynamic sparse mask for item ${i,j}$ of the attention matrix. Here, $c$ is a large constant, such as 1e5, ensuring that the less important attention weights for which $M_{i,j} = 0$ have values approaching zero after the softmax, i.e., $A_{i,j} \approx 0$.

The goal of the dynamic sparse attention system is to achieve greater speedup with minimal overhead while retaining as much of the attention weights as possible. Formally, this can be expressed as:
\begin{equation}
    \begin{aligned}
        \min \enspace & \enspace \enspace |\bm{A}(\bm{M}) - \bm{A}_{\text{dense}}|, \\
        \min \enspace &  t_{\text{sparse}}(\bm{M}) + t_{\text{overhead}}(\bm{M}),
    \end{aligned}
\end{equation}
where $t_{\text{sparse}}$ and $t_{\text{overhead}}$ represent the time spent on dynamic sparse attention computation and estimation of the approximate dynamic sparse pattern, respectively.

\restylefloat{algorithm}

\begin{wrapfigure}{r}{0.42\columnwidth}
\vspace{-20pt}
\begin{minipage}{0.42\textwidth}
\begin{algorithm}[H]
\captionsetup[algorithm]{singlelinecheck=off}
\caption{Kernel-Aware Sparse Pattern Search}
\label{alg:search}
\begin{algorithmic}
  \STATE {\bfseries Input:} $\boldsymbol{Q},\boldsymbol{K},\boldsymbol{V} \in \mathbb{R}^{S \times d_h}$, patterns $p$, search space $\rho$, target FLOPs $t$, initialized search space $\sigma$

    \LineComment{Build kernel-aware search space}
    \FOR{$i \gets 1$ to $|\sigma|$}
        \STATE $t_i \gets \text{FLOPs\_in\_kernel}(\sigma_i)$
        \WHILE{$|t_i - t| >\epsilon$}
        \STATE $\sigma_i \gets \text{ChangeSpace}(\sigma_i, p_i)$
        \STATE $t_i \gets \text{FLOPs\_in\_kernel}(\sigma_i)$
        \ENDWHILE
        \STATE $\rho \gets \rho \cup \sigma_i$
    \ENDFOR
    
    \LineComment{Search for optimal head pattern}
    \STATE $p_{\text{best}} \gets \phi$
    \STATE $\bm{y} \gets \text{Softmax}(\bm{Q}\bm{K}^\top/\sqrt{d})$
    \FOR{$i \gets 1$ to $|\rho|$}
        \STATE $\bm{y}_i \gets \text{SparseAttention}(\bm{Q}\bm{K}^\top/\sqrt{d}, \rho_i)$
        \STATE $p_{\text{best}} \gets \text{argmin}(\bm{y}_i - \bm{y}, p_{\text{best}})$
    \ENDFOR
    \STATE $\mathrm{return}\,\,\,p_{\text{best}}$
   
\end{algorithmic}
\end{algorithm}
\end{minipage}
\vspace{-10pt}
\end{wrapfigure}

\subsection{Speedup of Long-context LLM Inference via Dynamic Sparse Attention}
\label{3.2}
\paragraph{Kernel-Aware Optimal Sparse Pattern Search}

To achieve the best accuracy with limited FLOPs budget, we propose an offline Kernel-Aware Optimal Sparse Pattern Search method. In this step, we determine which sparse pattern will be used for each attention head, and the optimal setting for the pattern in real calculation (e.g., the number of vertical/slash lines in \textit{VS} pattern; or the number of top-k blocks in \textit{BS} patterns). As shown in Algorithm~\ref{alg:search}, we first create the search space based on a target FLOPs for each pattern, ensuring all potential candidates (i.e., different patterns with different settings) have similar computational cost. \textit{Kernel-aware} here indicates the computational cost reflects the real FLOPs in GPU kernels, instead of conceptual estimations, which is crucial to achieve the optimal acceleration.

\restylefloat{algorithm}

\begin{figure}[b]
\centering
\vspace{-20pt}
\begin{minipage}[t]{0.45\textwidth}
\vspace{0pt}
\centering
\begin{algorithm}[H]
\captionsetup[algorithm]{singlelinecheck=off}
\caption{Vertical-Slash Head}
\label{alg:vs_attention}
\begin{algorithmic}
  \STATE {\bfseries Input:} $\boldsymbol{Q},\boldsymbol{K},\boldsymbol{V} \in \mathbb{R}^{S \times d_h}$, $k_v,k_s \in \mathbb{N}$

    \LineComment{Approximate vertical and slash pattern (last\_q = 64)}
    \STATE $\boldsymbol{\hat{A}} \gets \mathrm{softmax}\left(\boldsymbol{Q}_{[-\text{last\_q}:]} \bm{K}^{\top} / \sqrt{d} + \bm{m}_{\text{casual}} \right)$
    
    \LineComment{Indices of top $k_v$ vertical line, sum in vertical}
    \STATE $\boldsymbol{i}_v \gets \mathrm{argtopk}\left(\mathrm{sum}_v(\boldsymbol{\hat{A}}), k_v\right)$

    \LineComment{Indices of top $k_s$ slash line, sum in slash}
    \STATE $\boldsymbol{i}_s \gets \mathrm{argtopk}\left(\mathrm{sum}_s(\boldsymbol{\hat{A}}), k_s\right)$
    
    \LineComment{Build sparse attention index}
    \STATE $\boldsymbol{i}_{vs} \gets \mathrm{sparseformat}(\boldsymbol{i}_v, \boldsymbol{i}_s)$

    \LineComment{Final dynamic sparse attention scores (only index block)}
    \STATE $\boldsymbol{A} \gets \mathrm{softmax}\left(\mathrm{sparse}(\boldsymbol{Q} \bm{K}^\top, \boldsymbol{i}_{vs}) / \sqrt{d}\right)$
           
    \LineComment{Sparse mixed scores and values}
    \STATE $\boldsymbol{y} \gets \mathrm{sparse}(\boldsymbol{A} \bm{V}, \boldsymbol{i}_{vs})$\\
    \STATE $\mathrm{return}\,\,\,\boldsymbol{y}$
   
\end{algorithmic}
\end{algorithm}
\end{minipage}
\hfill
\begin{minipage}[t]{0.45\textwidth}
\vspace{0pt}
\centering
\begin{algorithm}[H]
\captionsetup[algorithm]{singlelinecheck=off}
\caption{Block-Sparse Head}
\label{alg:block_sparse_attention}
\begin{algorithmic}
  \STATE {\bfseries Input:} $\boldsymbol{Q},\boldsymbol{K},\boldsymbol{V} \in \mathbb{R}^{S \times d_h}$, $k_b \in \mathbb{N}$

    \LineComment{Approximate block-sparse pattern (block\_size = 64)}
    \STATE $\boldsymbol{\hat{Q}} \gets \mathrm{MeanPooling}(\bm{Q},block\_size)$
    \STATE $\hat{\bm{K}} \gets \mathrm{MeanPooling}(\bm{K},block\_size)$
    \STATE $\boldsymbol{\hat{A}} \gets \mathrm{softmax}\left(\boldsymbol{\hat{Q}} \boldsymbol{\hat{K}} ^{\top} / \sqrt{d} + \bm{m}_{\text{casual}} \right)$
    
    \LineComment{Indices of top $k$ blocks}
    \STATE $\boldsymbol{i}_b \gets \mathrm{argtopk}\left(\boldsymbol{\hat{A}}, k_b\right)$
    
    \LineComment{Build sparse attention index}
    \STATE $\boldsymbol{i}_{b} \gets \mathrm{sparseformat}(\boldsymbol{i}_b)$

    \LineComment{Final dynamic sparse attention scores (only index block)}
    \STATE $\boldsymbol{A} \gets \mathrm{softmax}\left(\mathrm{sparse}(\boldsymbol{Q} \bm{K}^\top, \boldsymbol{i}_{b}) / \sqrt{d}\right)$
           
    \LineComment{Sparse mixed scores and values}
    \STATE $\boldsymbol{y} \gets \mathrm{sparse}(\boldsymbol{A} \bm{V}, \boldsymbol{i}_{b})$\\
    \STATE $\mathrm{return}\,\,\,\boldsymbol{y}$
   
\end{algorithmic}
\end{algorithm}
\end{minipage}
\vspace{-10pt}
\end{figure}

Next, we go through the search space with a reference example to decide the optimal pattern and setting. Specifically, we use recall of the attention output as the objective criterion when searching for the best pattern. This approach leverages FlashAttention~\cite{dao2023flashattention} to reduce GPU memory overhead and incorporates the information from the $\bm{V}$ matrix, enabling end-to-end selection of the best pattern, which further enhances performance.

\paragraph{Sparsity Indices Approximation and Dynamic Sparse Attention Calculation} 
During the inference stage, we will perform an online estimation on the attention matrix to dynamically determine the spatial distribution our sparse indices, based on the assigned patterns and the exact input.
After that, we conduct the sparse attention computations with our optimized GPU kernels. The implementation details of our kernels can be found in Appendix~\ref{subsec:kernel}. Noted that the sparse mask is static for \textit{A-shape} heads, so there is no overhead in building the dynamic masks, and only sparse calculation is required.

\textit{(i) \textit{Vertical-Slash} head.}
As shown in Algorithm~\ref{alg:vs_attention}, due to the continuity of vertical and slash lines, we matmul the last query vector $\bm{Q}_{[-\text{last\_q}:]}$ and key vector $\bm{K}$ to produce the estimated attention matrix $\boldsymbol{\hat{A}}$, which, in turn, is used to determine the indices for the vertical $\bm{i}_v$ and slash $\bm{i}_s$ lines.
After obtaining the sparse indices for the vertical and slash lines, we convert them into a sparse format $\bm{i}_{vs}$. Using these sparse indices, we perform block-sparse calculations of the attention weights and attention output.

\textit{(ii) \textit{Block-Sparse} head.}
Per Algorithm~\ref{alg:block_sparse_attention}, mean pooling is applied on $\bm{Q}$ and $\bm{K}$ to obtain \( \bm{\hat{Q}} \) and \( \bm{\hat{K}} \), respectively. The two matrices are multiplied to get the estimated block-level attention weights $\bm{\hat{A}}$. Since the mean pooling and matrix multiplication operations are commutative, the resulting attention weights are approximately equivalent to the actual attention weights after mean pooling. This allows us to approximate the actual attention weights' block-sparse pattern with minimal overhead. Similarly, we build a sparse index $\bm{i}_b$ and use it to compute the sparse attention weights and attention output.

\section{Experiments}
\label{sec:experiments}

In this section, we investigate two questions:
\textbf{(i) How effective is \method{}?} We evaluate our method on three general long-context benchmarks: InfiniteBench~\cite{zhang2024inftybench}, RULER~\cite{hsieh2024ruler}, and the Needle In A Haystack task~\cite{kamradt2023needle}, as well as the long-context language modeling task~\cite{rae2019compressive}. These benchmarks cover long-context QA, multi-hop QA, math reasoning, aggregation tasks, summarization, retrieval tasks, and code debugging, allowing us to assess \method{}'s effectiveness across a wide range of long-context scenarios.
\textbf{(ii) How efficient is \method{}?}
We delve into the end-to-end latency and its breakdown to evaluate the efficiency of \method{}.
Additional experimental, latency results, and analysis can be found in Appendix~\ref{sec:additional_experiment}, \ref{sec:pattern}, and \ref{sec:sparsity}.

\paragraph{Implementation Details}
Our experiments use four state-of-the-art long-context LLMs: LLaMA-3-8B-Instruct-262k\footnote{https://huggingface.co/gradientai/Llama-3-8B-Instruct-Gradient-262k}, LLaMA-3-8B-Instruct-1048k\footnote{https://huggingface.co/gradientai/Llama-3-8B-Instruct-Gradient-1048k}, GLM-4-9B-1M~\cite{glm2024chatglm}, and Yi-9B-200K~\cite{young2024yi}. Additionally, we tested Needle in A Haystack~\cite{kamradt2023needle} on Phi-3-Mini-128K~\cite{Abdin2024Phi3TR} and Qwen2-7B-128K~\cite{qwen}, as detailed in Appendix~\ref{subsec:needle}.
To guarantee stable results, we use greedy decoding in all experiments.
We provide a simple custom implementation of our method in PyTorch, built on FlashAttention~\cite{dao2023flashattention}, Triton~\cite{tillet2019triton}, and the dynamic sparse compiler PIT~\cite{zheng2023pit}. 
We set the target FLOPs $t$ to 1k global tokens and 4k local windows in the \textit{A-shape} pattern. We set $\text{last\_q}=64$ and $block\_size=64$ in the \textit{Vertical-Slash} and \textit{Block-Sparse} patterns, respectively.
The latency experiments are conducted on a single Nvidia A100 GPU in the bfloat16 format. More details are provided in Appendix~\ref{subsec:additional_implementation}.

\paragraph{Dataset \& Evaluation Metrics}
We use the provided metrics and scripts from the following benchmarks for evaluation. More details about dataset can be found in Appendix~\ref{dataset_detail}.

(i) InfiniteBench~\cite{zhang2024inftybench}: This benchmark consists of %
10 tasks, including retrieval tasks such as PassKey retrieval, Number retrieval, and KV retrieval, as well as representative realistic tasks like question-answering, coding, dialogue, and summarization. The average context length of InfiniteBench is about 214K tokens.

(ii) RULER~\cite{hsieh2024ruler}: A challenging long-context benchmark consisting of 4 categories and 13 complex tasks, including retrieval, multi-hop tracing and aggregation, and QA tasks. It contains subsets with different prompt lengths up to 128k tokens, allowing us to determine the actual context window size of the model based on the results. 

(iii) Needle In A Haystack~\cite{kamradt2023needle}: A long-context retrieval benchmark testing LLMs' performance with context window sizes up to 1M tokens where information placed at various positions.

(iv) PG-19~\cite{rae2019compressive}: Following StreamingLLM~\cite{xiao2023streamingllm} and H2O~\cite{zhang2023h2o}, we use PG-19 for long-context language modeling tasks with prompts up to 100k tokens.

\begin{table}[tb]
    \small
    \centering
    \setlength{\tabcolsep}{0.5mm}
    \vspace{-2ex}
    \caption{Performance of different methods with different base models on InfiniteBench~\cite{zhang2024inftybench}.}
    \resizebox{\columnwidth}{!}{
    \begin{tabular}{l|cccccccccc|c}
    \toprule
        Methods &  En.Sum & En.QA & En.MC & En.Dia & Zh.QA & Code.Debug & Math.Find & Retr.PassKey & Retr.Num & Retr.KV &  Avg. \\
    \midrule
    \textit{LLaMA-3-8B-262K} & 20.2 & 12.4 & 67.3 & 6.0 & 12.9 & 22.1 & 26.6 & 100.0 & 100.0 & 14.4 & 38.2 \\
    StreamingLLM & 21.0 & 8.2 & 40.2 & 10.0 & 10.4 & \textbf{25.9} & 30.0 & 86.8 & 5.1 & 0.8 & 23.8 \\
    StreamingLLM w/ dilated & 20.1 & 9.4 & 44.5 & \textbf{15.5} & 11.2 & 20.5 & 27.5 & 5.0 & 87.5 & 0.5 & 24.2 \\
    StreamingLLM w/ strided & 17.3 & 8.2 & 27.5 & 14.5 & 11.2 & 19.5 & 27.5 & 4.0 & 2.1 & 1.0 & 13.3 \\
    InfLLM & \textbf{24.1} & 7.8 & 45.0 & 6.0 & 11.4 & 19.5 & 32.9 & \textbf{100.0} & \textbf{100.0} & 1.2 & 34.8 \\
    Ours w/ static & 19.9 & 8.6 & 43.2 & 3.5 & 8.9 & 20.6 & 25.1 & 92.4 & 96.3 & 0.2 & 31.9 \\
    {\cellcolor[rgb]{0.925,0.957,1}}\textbf{Ours} & {\cellcolor[rgb]{0.925,0.957,1}}20.5 & {\cellcolor[rgb]{0.925,0.957,1}}\textbf{12.9} & {\cellcolor[rgb]{0.925,0.957,1}}\textbf{65.9} & {\cellcolor[rgb]{0.925,0.957,1}}7.5 & {\cellcolor[rgb]{0.925,0.957,1}}\textbf{12.5} & {\cellcolor[rgb]{0.925,0.957,1}}22.3 & {\cellcolor[rgb]{0.925,0.957,1}}\textbf{33.1} & {\cellcolor[rgb]{0.925,0.957,1}}\textbf{100.0} & {\cellcolor[rgb]{0.925,0.957,1}}\textbf{100.0} & {\cellcolor[rgb]{0.925,0.957,1}}\textbf{12.8} & {\cellcolor[rgb]{0.925,0.957,1}}\textbf{38.8} \\
    \midrule
    \textit{Yi-9B-200K} & 8.2 & 10.6 & 64.2 & 1.0 & 17.3 & 21.3 & 23.4 & 99.8 & 100.0 & 28.8 & 37.5 \\
    StreamingLLM & 5.4 & \textbf{14.2} & 38.0 & \textbf{4.0} & 18.8 & 18.8 & 22.3 & 39.2 & 6.1 & 1.6 & 16.8\\
    StreamingLLM w/ dilated & 5.7 & 4.2 & 15.0 & 0.0 & 18.2 & 0.0 & 2.9 & 0.0 & 0.0 & 0.0 & 4.2 \\
    StreamingLLM w/ strided & 6.1 & 4.5 & 9.8 & 0.0 & 16.9 & 0.0 & 3.1 & 1.5 & 0.0 & 0.0 & 4.6 \\
    InfLLM & 6.3 & 13.0 & 45.9 & 2.5 & \textbf{21.5} & 20.6 & 34.6 & 85.3 & 88.1 & 1.4 & 31.9\\
    Ours w/ static & 5.8 & 12.6 & 48.5 & 3.0 & 12.6 & 20.8 & \textbf{25.1} & 60.9 & 38.5 & 1.0 & 22.9 \\
    {\cellcolor[rgb]{0.925,0.957,1}}\textbf{Ours} & {\cellcolor[rgb]{0.925,0.957,1}}\textbf{7.9} & {\cellcolor[rgb]{0.925,0.957,1}}11.2 & {\cellcolor[rgb]{0.925,0.957,1}}\textbf{64.2} & {\cellcolor[rgb]{0.925,0.957,1}}1.0 & {\cellcolor[rgb]{0.925,0.957,1}}17.9 & {\cellcolor[rgb]{0.925,0.957,1}}\textbf{24.1} & {\cellcolor[rgb]{0.925,0.957,1}}23.1 & {\cellcolor[rgb]{0.925,0.957,1}}\textbf{99.5} & {\cellcolor[rgb]{0.925,0.957,1}}\textbf{100.0} & {\cellcolor[rgb]{0.925,0.957,1}}\textbf{27.6} & {\cellcolor[rgb]{0.925,0.957,1}}\textbf{37.7} \\
    \midrule
    \textit{GLM-4-9B-1M} & 28.3 & 9.7 & 68.6 & 39.5 & 12.1 & 29.4 & 38.9 & 100.0 & 100.0 & 41.0 & 46.7 \\
    StreamingLLM & 27.7 & 6.4 & 40.2 & 12.5 & 10.8 & 27.7 & 21.1 & 97.1 & 25.6 & 0.6 & 27.0\\
    InfLLM & 28.0 & 7.3 & 45.0 & 14.0 & 10.7 & 27.9 & \textbf{39.4} & 98.0 & \textbf{100.0} & 2.6 & 37.3\\
    {\cellcolor[rgb]{0.925,0.957,1}}\textbf{Ours} & {\cellcolor[rgb]{0.925,0.957,1}}\textbf{28.8} & {\cellcolor[rgb]{0.925,0.957,1}}\textbf{9.6} & {\cellcolor[rgb]{0.925,0.957,1}}\textbf{68.6} & {\cellcolor[rgb]{0.925,0.957,1}}\textbf{38.5} & 
    {\cellcolor[rgb]{0.925,0.957,1}}\textbf{12.0} & {\cellcolor[rgb]{0.925,0.957,1}}\textbf{30.7} & {\cellcolor[rgb]{0.925,0.957,1}}39.1 & {\cellcolor[rgb]{0.925,0.957,1}}\textbf{100.0} & {\cellcolor[rgb]{0.925,0.957,1}}\textbf{100.0} & {\cellcolor[rgb]{0.925,0.957,1}}\textbf{43.0} & {\cellcolor[rgb]{0.925,0.957,1}}\textbf{47.0} \\
    \bottomrule
    \end{tabular}
    }
    \label{tab:main_results_infinitebench}
\end{table}

\paragraph{Baselines}
We include five training-free sparse attention approaches as our baselines:
1) StreamingLLM~\cite{xiao2023streamingllm}, which corresponds to the \textit{A-shape} pattern. We use 1k global tokens and 4k local windows in all our experiments;
2) StreamingLLM w/ dilated~\cite{beltagy2020longformer}, which sets dilated local windows with intervals in the local windows direction. We use 1k global tokens and 8k dilated attention windows with an interval of 1;
3) StreamingLLM w/ strided~\cite{child2019sparsetransformers}, which retains local windows while adding dilated attention. We use 1k global tokens, 2k local windows, and 4k dilated attention windows with an interval of 1;
4) InfLLM~\cite{xiao2024infllm}, which uses a memory unit to process streaming long sequences. Following the paper, we set 128 global tokens and 8k local windows in all experiments;
5) Ours w/ static, which utilizes static sparse indices in the \textit{Vertical-Slash} and \textit{Block-Sparse} heads.
For all baselines, we perform sparse computation only during the pre-filling stage, while retaining dense computation during the decoding stage.

\paragraph{InfiniteBench}

\begin{table}[!b]
\centering
\caption{Performance (\%) of different models and different methods on RULER~\cite{hsieh2024ruler} evaluated at lengths from 4k to 128k.}
\resizebox{0.84\columnwidth}{!}{
\begin{tabular}{l|cccccccc|c}
\toprule
Methods & Claimed & Effective & 4K & 8K & 16K & 32K & 64K & 128K & Avg. \\ \midrule
\textit{LLaMA-3-8B-262K} & 262K & 16K & 97.2 & 91.8 & 87.3 & 80.8 & 77.4 & 72.2 & 84.4 \\ 
StreamingLLM & - & 4K & 97.2 & 38.1 & 37.5 & 17.2 & 14.2 & 9.4 & 35.0 \\
StreamingLLM w/ dilated & - & \textless 4K & 23.4 & 0.7 & 1.4 & 18.8 & 16.5 & 15.6 & 12.7 \\ 
StreamingLLM w/ strided & - & \textless 4K & 2.0 & 0.7 & 0.6 & 0.6 & 0.7 & 1.3 & 1.0 \\
InfLLM & - & 4K & 89.4 & 79.8 & 70.1 & 55.6 & 43.0 & 39.5 & 62.9 \\
{\cellcolor[rgb]{0.925,0.957,1}}\textbf{Ours} & {\cellcolor[rgb]{0.925,0.957,1}}- & {\cellcolor[rgb]{0.925,0.957,1}}\textbf{32K} &
{\cellcolor[rgb]{0.925,0.957,1}}\textbf{97.7} & {\cellcolor[rgb]{0.925,0.957,1}}\textbf{91.2} & {\cellcolor[rgb]{0.925,0.957,1}}\textbf{88.5} & {\cellcolor[rgb]{0.925,0.957,1}}\textbf{85.0} & {\cellcolor[rgb]{0.925,0.957,1}}\textbf{82.3} & {\cellcolor[rgb]{0.925,0.957,1}}\textbf{77.6} & {\cellcolor[rgb]{0.925,0.957,1}}\textbf{87.0} \\
\midrule
\textit{Yi-9B-200K} & 200K & 8K & 91.9 & 90.2 & 78.8 & 76.3 & 68.1 & 62.9 & 78.1 \\
StreamingLLM & - & 4K & 91.9 & 37.8 & 33.9 & 18.6 & 13.0 & 12.8 & 34.3 \\
StreamingLLM w/ dilated & - & \textless
4K & 44.8 & 42.8 & 38.5 & 29.8 & 26.8 & 23.9 & 34.4 \\
StreamingLLM w/ strided & - &  \textless 4K & 2.6 & 0.7 & 0.6 & 0.6 & 1.2 & 0.5 & 1.1 \\
InfLLM & - &  \textless 4K  & 80.3 & 83.9 & 60.7 & 45.2 & 38.6 & 30.2 & 56.5 \\
{\cellcolor[rgb]{0.925,0.957,1}}\textbf{Ours} & 
{\cellcolor[rgb]{0.925,0.957,1}}- & 
{\cellcolor[rgb]{0.925,0.957,1}}8K & 
{\cellcolor[rgb]{0.925,0.957,1}}\textbf{92.3} & 
{\cellcolor[rgb]{0.925,0.957,1}}\textbf{89.7} & 
{\cellcolor[rgb]{0.925,0.957,1}}\textbf{79.0} &
{\cellcolor[rgb]{0.925,0.957,1}}\textbf{73.8} &
{\cellcolor[rgb]{0.925,0.957,1}}\textbf{64.7} &
{\cellcolor[rgb]{0.925,0.957,1}}\textbf{56.9} &
{\cellcolor[rgb]{0.925,0.957,1}}\textbf{74.7}\\
\midrule
\textit{GLM-4-9B-1M} & 1M & 64K & 93.8 & 91.6 & 89.3 & 87.4 & 85.2 & 80.8 & 88.0 \\
StreamingLLM & - & 4K & 93.8 & 66.9 & 58.5 & 51.4 & 45.9 & 39.1  & 59.3 \\
InfLLM & - &  8K  & \textbf{94.7} & 89.5 & 76.4 & 66.5 & 56.8 & 53.5 & 72.9 \\
{\cellcolor[rgb]{0.925,0.957,1}}\textbf{Ours} & 
{\cellcolor[rgb]{0.925,0.957,1}}- & 
{\cellcolor[rgb]{0.925,0.957,1}}64K & 
{\cellcolor[rgb]{0.925,0.957,1}}94.6 & {\cellcolor[rgb]{0.925,0.957,1}}\textbf{93.1} & {\cellcolor[rgb]{0.925,0.957,1}}\textbf{91.0} & {\cellcolor[rgb]{0.925,0.957,1}}\textbf{89.6} & {\cellcolor[rgb]{0.925,0.957,1}}\textbf{85.5} & {\cellcolor[rgb]{0.925,0.957,1}}\textbf{84.0} & {\cellcolor[rgb]{0.925,0.957,1}}\textbf{89.6}\\
\bottomrule
\end{tabular}}
\label{tab:ruler_results}
\end{table}

As shown in Table~\ref{tab:main_results_infinitebench}, \method{} achieves the best overall performance on InfiniteBench compared to baseline methods. Remarkably, \method{} matches or even slightly surpasses the performance of the original full attention baseline on some tasks, despite the significant acceleration it provided. From the perspective of different tasks, our method not only performs well in natural language tasks such as summarization, QA, and code, but also maintains the original model's performance on retrieval-related tasks. Baseline methods such as StreamingLLM, on the contrary, struggle with these retrieval tasks. Additionally, on tasks such as dialogue QA, using local attention mechanisms can better handle these tasks, while our performance is closer to the original results, indicating that our method is not solely based on local windows. Extending the local windows' intervals in StreamingLLM, i.e., w/ dilated and w/ strided, has minimal impact on the model’s performance.

\paragraph{RULER}
To further reveal the true potential of our method in long-context LLMs, we evaluate \method{} with the state-of-the-art long-context challenge, RULER. As shown in Table~\ref{tab:ruler_results}, \method{} effectively maintains the long-context performance even in complex multi-hop or aggregation tasks in RULER. It even outperforms the original full attention for testing lengths beyond 32K, achieving effective context windows of 32K and 64K (context with performance over 85\% is considered effective~\cite{hsieh2024ruler}) in LLaMA-3-8B-262K and GLM-4-9B-1M.

\begin{figure*}[htb]
  \centering
  \subfloat[LLaMA-3-8B-Instruct-262K]{
    \label{sfig:ppl_llama3}
    \includegraphics[width=0.5\columnwidth]{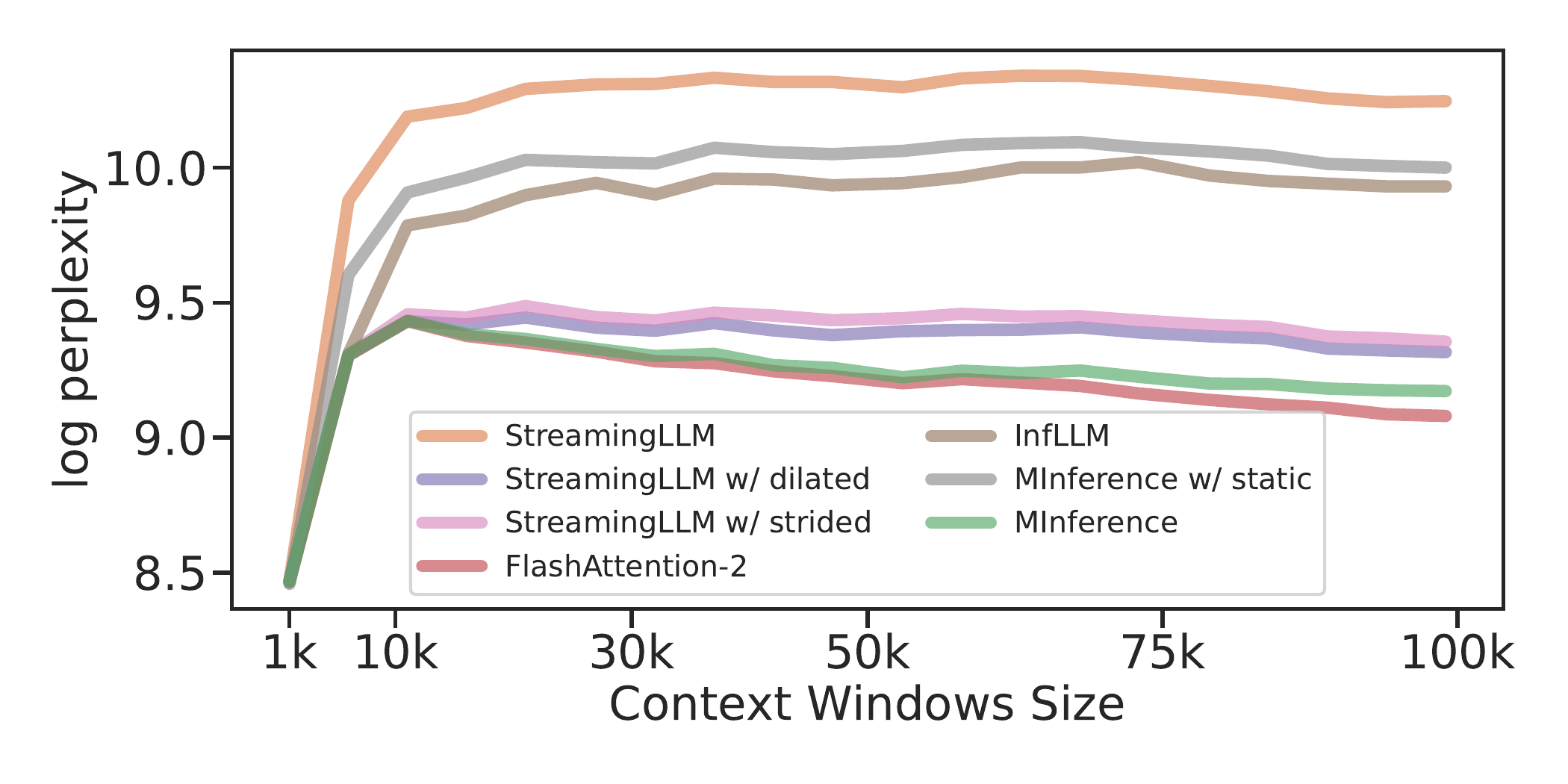}}
  \subfloat[Yi-9B-200K]{
    \label{sfig:ppl_yi}
    \includegraphics[width=0.5\columnwidth]{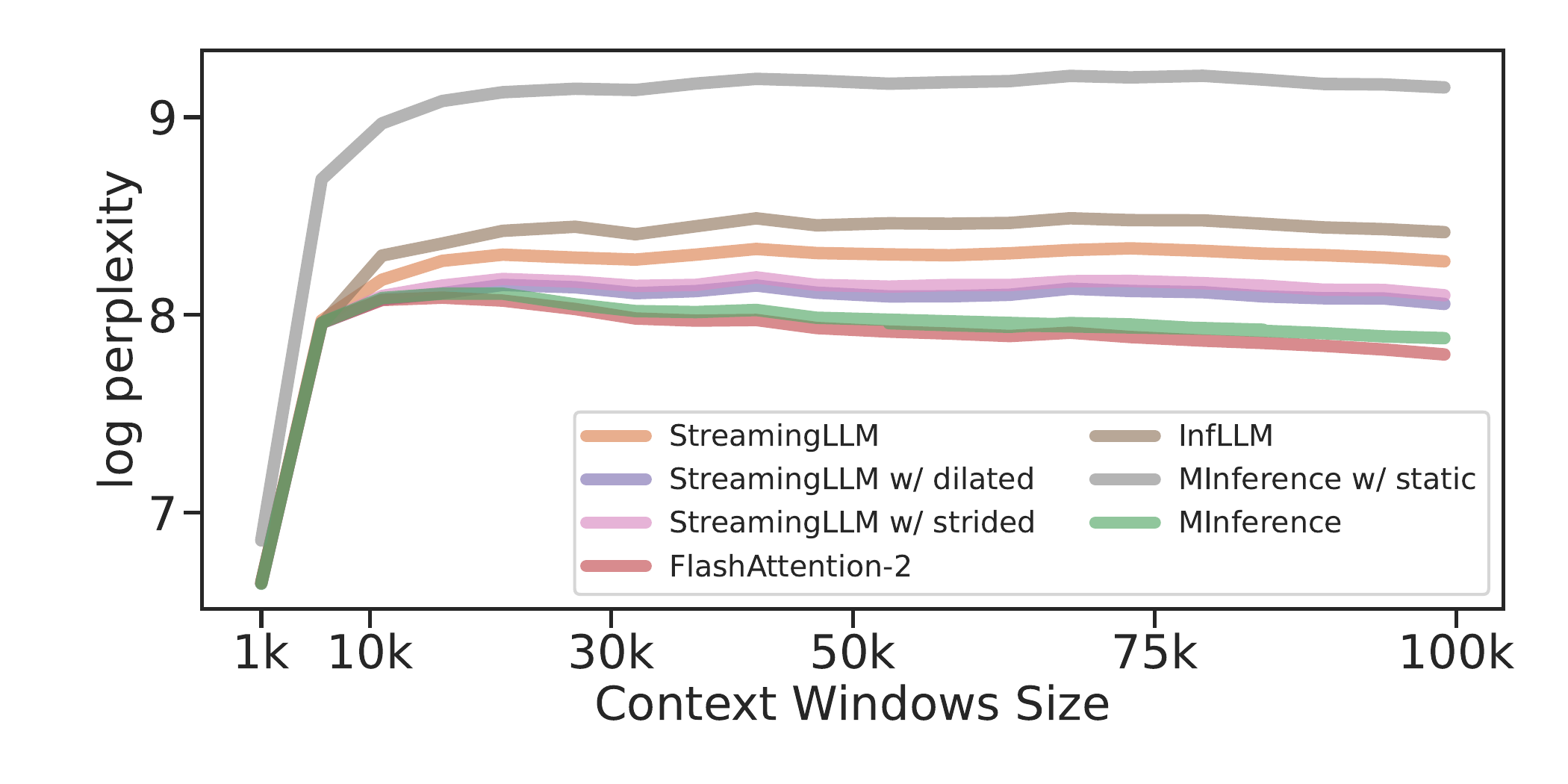}}
  \caption{Perplexity results on PG-19~\cite{rae2019compressive} using different models and methods.}
  \label{fig:ppl}
\end{figure*}

\paragraph{Language Modeling}
Following the approach of StreamingLLM~\cite{xiao2023streamingllm} and H2O~\cite{zhang2023h2o}, we evaluate our methods against baselines on the language modeling task based on the PG-19 dataset~\cite{rae2019compressive}. As shown in \ref{fig:ppl}, our method yields best results compared to other sparse approaches, and exhibits minimal divergence compared to the full attention baseline. For prompts of 100K token, our perplexity is only 0.2 higher than the full attention, but lower than StreamingLLM for 0.25 and 0.75 on the Yi-9B-200K and LLaMA-3-262K models respectively.

\begin{wrapfigure}{r}{0.5\columnwidth}
    \vspace{-12pt}
    \centering
    \includegraphics[width=1\linewidth]{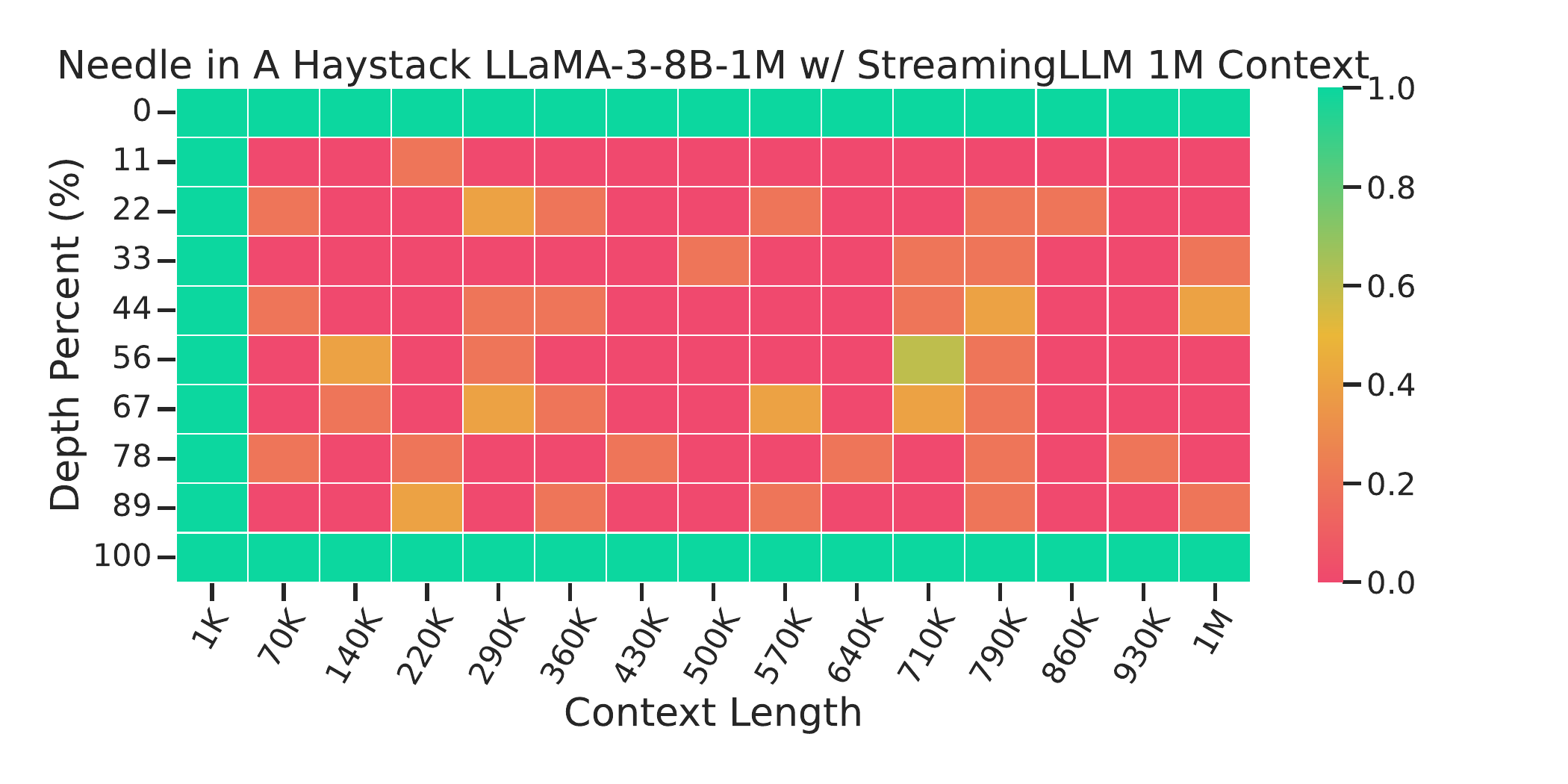}
    \caption{Results on Needle In A Haystack of StreamingLLM~\cite{xiao2023streamingllm} in LLaMA-3-8B-1M.}
    \label{fig:needle_streamllm}
    \vspace{-12pt}
\end{wrapfigure}

\paragraph{Needle In A Haystack}

Comparing Fig.~\ref{sfig:needle_ours_result} to Fig.~\ref{fig:needle_streamllm}, our method effectively retains the ability to process information at different positions across various context windows, ranging from 1k to 1M tokens. In contrast, methods like StreamingLLM and InfLLM (as shown in Appendix~\ref{subsec:needle}), while effective at reducing latency, experience a sharp decline in performance once critical information extends beyond the range of global tokens and local windows.

\paragraph{Ablation Study}
To evaluate the contributions of different components in \method{}, we introduce four variants for the ablation study:
(1) Ours w/ static, which uses a static sparse mask in the \textit{Vertical-Slash} and \textit{Block-Sparse} patterns;
(2) Ours w/ only A-shape, which is equivalent to StreamingLLM;
(3) Ours w/ only block-sparse, which uses only the \textit{Block-Sparse} pattern in the dynamic sparse calculation.
(4) Ours w/ only vertical-slash, which uses only the \textit{Vertical-Slash} pattern in the dynamic sparse calculation.

\begin{table}[htb]
    \small
    \centering
    \setlength{\tabcolsep}{0.5mm}
    \vspace{-2ex}
    \caption{Performance of different ablation methods using LLaMA-3-8B-Instruct-262K on InfiniteBench~\cite{zhang2024inftybench}.}
    \resizebox{\columnwidth}{!}{
    \begin{tabular}{l|cccccccccc|c}
    \toprule
        Methods &  En.Sum & En.QA & En.MC & En.Dia & Zh.QA & Code.Debug & Math.Find & Retr.PassKey & Retr.Num & Retr.KV &  Avg. \\
    \midrule
    
    Ours & 20.5 & 12.9 & 65.9 & 7.5 & 12.5 & 22.3 & 33.1 & 100.0 & 100.0 & 12.8 & 38.8 \\
    Ours w/ only block-sparse & 12.4 & 3.4 & 5.7 & 6.0 & 3.1 & 12.2 & 24.0 & 59.5 & 60.3 & 0.0 & 18.7 \\
    Ours w/ only vertical-slash & 19.6 & 12.0 & 62.1 & 9.5 & 11.7 & 21.6 & 29.1 & 100.0 & 100.0 & 5.0 & 37.1 \\
    \bottomrule
    \end{tabular}
    }
    \label{tab:ablation}
\end{table}

Tables~\ref{tab:main_results_infinitebench}, \ref{tab:ruler_results}, and \ref{tab:ablation} present the ablation results. It first proves that using static indices significantly degrades LLM performance, especially in highly dynamic tasks like KV retrieval, where accuracy nearly drops to zero. This highlight the necessity of our dynamic strategy and the effectiveness of our dynamically built sparse indices. Additionally, remove any pattern from the three leads to varying degrees of performance degradation. Specifically, "only A-shape" can only capture information within local windows. The "only block-sparse" variant using only the \textit{BS} pattern, also results in significant performance declines. On the other hand, "only vertical-slash" manages to preserve most of the performance due to its balance between dynamicity and the StreamingLLM pattern, but still fall behind the full version of our method.

\paragraph{Latency}

Fig.~\ref{sfig:speedup} and \ref{fig:latency_breakdown_detail} shows the latency and breakdown of \method{} across different context windows on a single A100. At 100K, 300K, 500K, and 1M tokens, our method achieves speedups of 1.8$\times$, 4.1$\times$, 6.8$\times$, and 10$\times$, respectively. It reduces the pre-filling latency from 30 mins to 3 mins on a single A100 for a prompt of 1M token. By further utilizing tensor parallel~\cite{nnscaler-osdi24} and context parallel~\cite{Liu2023RingAW,jacobs2023deepspeed}, this latency can be reduced to 22 seconds on 8x A100 GPUs. This significantly lowers the deployment cost of long-context LLMs and enhances user experience. 
And since our kernel is implemented based on Triton, it can be easily ported to other devices and achieve similar speedups, such as on the H100 or MI300X.
Additionally, analyzing the latency breakdown, we found about 5\%-20\% of the overhead is spent on dynamic sparse index building, while the remaining time is spent on dynamic sparse calculation.

\paragraph{Integrate with KV cache compression methods}

We also combined MInference with a state-of-the-art KV cache compression method SnapKV~\cite{li2024snapkv}, as shown in Table~\ref{tab:snapkv_results_infinitebench}. This proves our method is compatible with KV cache compression techniques. For most tasks, performance remains nearly unchanged, with the average score even showing a slight increase, which further demonstrates the potential practical value of our method as an optimization for serving long-context LLMs. This phenomenon is also observed in other works, such as ShadowKV~\cite{sun2024shadowkv}.
\begin{table}[htb]
    \small
    \centering
    \setlength{\tabcolsep}{0.5mm}
    \vspace{-2ex}
    \caption{Performance of different methods on InfiniteBench~\cite{zhang2024inftybench} using SnapKV~\cite{li2024snapkv} in the decoding stage.}
    \resizebox{\columnwidth}{!}{
    \begin{tabular}{l|cccccccccc|c}
    \toprule
        Methods &  En.Sum & En.QA & En.MC & En.Dia & Zh.QA & Code.Debug & Math.Find & Retr.PassKey & Retr.Num & Retr.KV &  Avg. \\
    \midrule
    
    LLaMA-3 w/ SnapKV & 18.0 & \textbf{11.8} & 65.5 & 2.5 & 12.0 & 21.3 & 26.6 & \textbf{100.0} & \textbf{100.0} & 1.8 & 36.0 \\
    \textbf{Ours} w/ SnapKV & \textbf{18.9} & 11.7 & \textbf{66.4} & \textbf{6.5} & \textbf{12.1} & \textbf{21.8} & \textbf{33.1} & \textbf{100.0} & \textbf{100.0} & \textbf{2.0} & \textbf{37.3} \\ 
    \bottomrule
    \end{tabular}
    }
    \label{tab:snapkv_results_infinitebench}
\end{table}

\paragraph{Scaling-up on Larger LLMs}

We also evaluated MInference on larger LLMs, such as LLaMA-3-70B-1M\footnote{https://huggingface.co/gradientai/Llama-3-70B-Instruct-Gradient-262k}. As shown in Table~\ref{tab:70B_results_infinitebench}, MInference maintains strong performance even in larger models. Notably, in dynamic tasks such as KV retrieval, MInference can match or even slightly improve performance compared to full attention. In contrast, baselines like InfLLM generally struggle with tasks such as KV retrieval.

\begin{table}[htb]
    \small
    \centering
    \setlength{\tabcolsep}{0.5mm}
    \vspace{-2ex}
    \caption{Performance of different methods using LLaMA-3-70B-Instruct-262K on InfiniteBench~\cite{zhang2024inftybench}.}
    \resizebox{\columnwidth}{!}{
    \begin{tabular}{l|cccccccccc|c}
    \toprule
        Methods &  En.Sum & En.QA & En.MC & En.Dia & Zh.QA & Code.Debug & Math.Find & Retr.PassKey & Retr.Num & Retr.KV &  Avg. \\
    \midrule
    
    \textit{LLaMA-3-70B-262K} & 20.7 & 10.3 & 84.2 & 9.5 & 14.0 & 33.2 & 61.7 & 97.0 & 100.0 & 34.0 & 46.5 \\
    StreamingLLM & 20.5 & 8.5 & 52.0 & \textbf{10.0} & 12.6 & 27.4 & 61.1 & 14.0 & 10.0 & 0.0 & 21.6 \\
    InfLLM & \textbf{24.1} & 8.1 & 57.0 & \textbf{10.0} & 12.9 & 27.4 & 52.3 & \textbf{100.0} & \textbf{100.0} & 0.0 & 39.2\\
    {\cellcolor[rgb]{0.925,0.957,1}}\textbf{Ours} & {\cellcolor[rgb]{0.925,0.957,1}}20.6 & {\cellcolor[rgb]{0.925,0.957,1}}\textbf{10.1} & {\cellcolor[rgb]{0.925,0.957,1}}\textbf{83.4} & {\cellcolor[rgb]{0.925,0.957,1}}\textbf{10.0} & {\cellcolor[rgb]{0.925,0.957,1}}\textbf{14.1} & {\cellcolor[rgb]{0.925,0.957,1}}\textbf{34.1} & {\cellcolor[rgb]{0.925,0.957,1}}\textbf{61.9} & {\cellcolor[rgb]{0.925,0.957,1}}\textbf{100.0} & {\cellcolor[rgb]{0.925,0.957,1}}\textbf{100.0} & {\cellcolor[rgb]{0.925,0.957,1}}\textbf{39.0} & {\cellcolor[rgb]{0.925,0.957,1}}\textbf{47.3} \\ 
    \bottomrule
    \end{tabular}
    }
    \label{tab:70B_results_infinitebench}
\end{table}

\section{Related Works}

\paragraph{Sparse Attention}

Due to the quadratic complexity of the attention mechanism, many previous works have focused on sparse attention to improve the efficiency of Transformers. These methods include static sparse patterns, cluster-based sparse approaches, and dynamic sparse attention. Static sparse patterns include techniques such as sliding windows~\cite{jiang2023mistral,Abdin2024Phi3TR}, dilated attention~\cite{child2019sparsetransformers,shi2021sparsebert,ding2023longnet}, and mixed sparse patterns~\cite{beltagy2020longformer,zaheer2020big,lagunas2021block}.
Cluster-based sparse methods include hash-based~\cite{kitaev2020reformer} and kNN-based~\cite{Roy2020EfficientCS, nawrot2024dynamic} methods. All of the above methods require pre-training the model from scratch, which makes them infeasible to be directly used as a plugin for reay-to-use LLMs.
Recently, there has been work~\cite{dao2024transformers,zimerman2024unified} to unify state space models~\cite{gu2021efficiently,gu2023mamba,dao2024transformers}, and linear attention~\cite{katharopoulos2020transformers,sun2023retentive} into structured masked attention.
Additionally, some works~\cite{wang2021spatten,liu2021transformer,ribar2023sparq} leverage the dynamic nature of attention to predict sparse patterns dynamically. However, these approaches often focus on low-rank hidden states during the dynamic pattern approximation or use post-statistical methods to obtain the sparse mask, which introduce substantial overhead in the estimation step, making them less useful for long-context LLMs.

\paragraph{Scaling Context Windows of LLMs}

Recent research has focused on expanding the context window of pre-trained LLMs, that enables LLMs to handle more complex real-life applications~\cite{Jimenez2023SWEbenchCL, Park2023GenerativeAI}. These methods can be categorized into:
1) Staged pre-training~\cite{nijkamp2023xgen7b,fu2024data};
2) Modifying or interpolating position embeddings~\cite{Press2021TrainST, Chen2023ExtendingCW, Peng2023YaRNEC, Ding2024LongRoPEEL};
3) Utilizing external memory modules for context storage~\cite{bertsch2023unlimiformer, tworkowski2023focused, xiao2024infllm};
4) Expanding computations across multiple devices in a distributed manner~\cite{Liu2023RingAW}.
However, these methods do not alleviate the high inference costs in long-context processing.

\paragraph{Long-Context LLM Inference}

Recent studies~\cite{fu2024challenges} have tackled the high computational cost of attention and substantial KV cache storage in long-context scenarios from two angles: pre-filling and decoding. Pre-filling optimizations are primarily categorized as State Space Models~\cite{gu2021efficiently,gu2023mamba}, linear attention methods~\cite{sun2023retentive,peng2023rwkv}, memory-based methods~\cite{munkhdalai2024leave,ho2024block}, hybrid methods~\cite{lieber2024jamba,ren2024samba}, and prompt compression methods~\cite{li2023compressing,jiang2023llmlingua,jiang2023longllmlingua,pan2024llmlingua}. However, these approaches require training from scratch or additional overhead and are difficult to implement directly in pretrained long-context LLMs.
Recently, some studies~\cite{iceformer,xiao2024infllm,liu2024retrievalattention} have focused on using kNN or cluster-based sparse attention to accelerate LLM inference. However, these methods often lead to reduced accuracy, limited speedup, or are restricted to CPU scenarios.

In contrast, optimizations for the decoding stage are divided into:
1) Reusing attention KV to reduce KV cache storage~\cite{shazeer2019fast,ainslie2023gqa,sun2024you,deepseekv2,nawrot2024dynamic};
2) Static KV cache dropping~\cite{xiao2023streamingllm,han2023lminfinite};
3) Dynamic KV cache dropping~\cite{zhang2023h2o,liu2023scissorhands,ge2023fastgen,oren2024transformers,li2024snapkv,anagnostidis2024dynamic};
4) Dynamic KV cache offloading~\cite{ribar2023sparq,dai2024sequence,tang2024quest,liu2024retrievalattention,chen2024magicpig,sun2024shadowkv};
5) Methods for restoring performance loss due to KV cache compression~\cite{adnan2024keyformer,dong2024less};
6) Hierarchical speculative decoding methods~\cite{sun2024triforce,chen2024magicdec};
7) KV cache quantitation~\cite{liu2024kivi}.
Nevertheless, these methods do not address the heavy computational burden of the attention in the pre-filling stage.

\section{Conclusion}

This paper addresses the expensive computational cost and the unacceptable latency of the attention calculations in the pre-filling stage of long-context LLMs. We propose \method{}, a method that accelerates the pre-filling stage by leveraging dynamic sparse attention with spatial aggregation patterns. Specifically, we categorize attention heads into three types: \textit{A-shape}, \textit{Vertical-Slash}, and \textit{Block-Sparse}. Using a kernel-aware optimal sparse pattern search method, we identify the optimal pattern for each head. Subsequently, we utilize a fast approximation approach to build dynamic sparse masks for different inputs, and then apply these mask to perform sparse attention calculations. Experimental results on benchmarks such as InfiniteBench, RULER, language modeling, and Needle In A Haystack demonstrate that our method effectively maintains the long-context capabilities of LLMs while achieving up to a $10\times$ speedup, reducing the latency from 30 minutes to 3 minutes per prompt for 1 million token prompts on a single A100 GPU.
Additionally, we have found that similar dynamic sparse attention patterns also exist in both multi-modal LLMs~\cite{wan2024lookmlookonceoptimizationkv} and encoder-decoder LLMs~\cite{raffel2020exploring}. 
Using MInference for pre-filling stage inference acceleration holds great promise.

{
\bibliographystyle{alpha}
\bibliography{ref}
}

\newpage
\appendix
\section{Limitations}
\label{sec:limitations}

As the context length decreases, the time required to build the dynamic index becomes more significant as attention computation time decreases. For example, with a 10k context, the time spent on building the index increases from 5\% to 30\%, resulting in overall end-to-end latency approaching that of FlashAttention. However, this overhead proportion gradually decreases as the prompt lengthens. Additionally, when using a higher sparsity rate, the model performance may noticeably decline.

\section{Broader Impacts}
\label{sec:impacts}

\method{} effectively accelerates the inference of long-context LLMs, facilitating their deployment and application. By enabling lower latency, it can reduce the deployment costs of LLMs, especially for long-context LLMs, helping to democratize access to advanced AI. It also promotes further research and development in related fields.

\section{Experiment Details}
\label{sec:appendix_experiments_details}

\subsection{Dataset Details}
\label{dataset_detail}

\paragraph{InfiniteBench~\cite{zhang2024inftybench}} includes 10 tasks designed to test various aspects of long-context processing. Specifically, these tasks cover entire novel summarization, open-form question answering based on novels, multiple-choice question answering on novels, question answering on long drama scripts, question answering on Chinese texts, debugging large code repositories, identifying the largest/smallest number in arrays, and retrieval tasks with varying pattern lengths. The average token length for these tasks is 214k, and they include 3,992 examples.

\paragraph{RULER~\cite{hsieh2024ruler}} is a recent synthetic benchmark suite for long-context evaluation with 13 complex tasks across four categories.
The retrieval category includes Single Needle-in-a-Haystack (S-NIAH), where a single key-value pair is inserted into noisy text, and the model must retrieve the value. Multi-keys Needle-in-a-Haystack (MK-NIAH) involves multiple keys, and the model retrieves one specific value among hard distractors. The Multi-values Needle-in-a-Haystack (MV-NIAH) task requires retrieving all values associated with a single key, while the Multi-queries Needle-in-a-Haystack (MQ-NIAH) task involves retrieving values for multiple keys. The Multi-hop Tracing category includes Variable Tracking (VT), where the model traces and returns all variable names pointing to the same value through variable bindings. The aggregation category introduces Common Words Extraction (CWE), where the model identifies the top-K common words from a mixture of common and uncommon words, and Frequent Words Extraction (FWE), where the model identifies the most frequent words from a Zeta distribution. The Question Answering (QA) category extends existing short-context QA datasets by adding distracting paragraphs, challenging the model to answer questions based on relevant information surrounded by distractors. These tasks provide a comprehensive evaluation of long-context modeling capabilities, covering multi-hop reasoning, aggregation, and complex question answering. Following \cite{hsieh2024ruler}, we test models on 4K, 8K, 16K, 32K, 64K, and 128K context lengths, including 2,600 examples per length.

\paragraph{Needle In A Haystack task~\cite{kamradt2023needle}} 
evaluates the performance of retrieval-augmented generation (RAG) systems by embedding specific, targeted information (the "needle") within a large, complex body of text (the "haystack"). The test assesses a language model's ability to identify and utilize this specific piece of information amidst a vast amount of data. Both RULER and the needle test iterate over various context lengths and document depths (where the ground-truth is placed in the prompt) to measure the long-context performance. 
Here we scale the Needle In A Haystack task to 1M context length, including 750 examples.

\paragraph{PG-19~\cite{rae2019compressive}} The perplexity on long text is also often used by researchers to evaluate the language modeling performance of long-context LLMs. PG-19 is a suitable test set for this task, as it includes texts as long as 500K tokens. Perplexity is used as the metric indicating how well a model predicts the next token in a sequence. Our experiments are conducted on 1,000 random samples from PG-19 that are longer than 100K tokens.

\subsection{Additional Implementation Details}
\label{subsec:additional_implementation}
Our experiments are based on a number of state-of-the-art long-context LLMs: 1) LLaMA-3-8B-Instruct-262k\footnote{https://huggingface.co/gradientai/Llama-3-70B-Instruct-Gradient-262k} is a LLaMA-3 variant with further NTK-aware interpolation and minimal fine-tuning with Ring Attention, which achieved SOTA results on long-context assessments such as the Needle In A Haystack test; 2) LLaMA-3-8B-Instruct-1048k\footnote{https://huggingface.co/gradientai/Llama-3-8B-Instruct-Gradient-1048k} is similar to LLaMA-3-8B-Instruct-262k, but supports context lengths up to 1M tokens; 3) Yi-9B-200K~\cite{young2024yi} is a SOTA LLM that balances long-context performance with general capabilities; 4) Phi-3-Mini-128K \cite{Abdin2024Phi3TR} a small but powerful language model that offers capabilities equivalent to models ten times its size with up to 128K context window powered by LongRoPE \cite{Ding2024LongRoPEEL}; 5) Qwen2-7B-128K \cite{qwen} is a recently release update of Qwen series model with up to 128K context window that achieve superior or comparable performance compared to LLaMA-3; 6) GLM-4-9B-1M \cite{glm2024chatglm} has been improved from its predecessor in terms of a 1M context window, performance on downstream tasks and inference efficiency. To guarantee stable results, we use greedy decoding in all tests. Our kernel implementations are developed and optimized based on the dynamic sparse compiler PIT \cite{zheng2023pit} in the Triton language \cite{tillet2019triton}. The latency experiments are done on a single Nvidia A100 GPU using bfloat16.
We provide a simple custom implementation of attention in PyTorch, building on FlashAttention and Triton.

We set the target FLOPs $t$ to be the same as 1k global tokens and 4k local window tokens in the \textit{A-shape} pattern.
The step size of $\text{ChangeSpace}$ is set to 50, with the corresponding search space shown in Table~\ref{tab:search_space}. Additionally, we use only one sample as our validation set from KV retrieval synthetic data with 30k token inputs, which exhibits strong generalization and stability across different lengths and domains. The search time is approximately 15 minutes on a single A100.
Additionally, we use the same optimal sparse pattern configuration for both the LLaMA-3-8B-Instruct-262K model and the LLaMA-3-8B-Instruct-1M model. The specific distribution is shown in Fig.~\ref{fig:pattern_distribution}.

\begin{table}[ht]
    \centering
    \caption{Kernal-aware optimal head pattern search space. In this context, \textit{A-shape} represents the global tokens and local window number, \textit{Vertical-Slash} represents the Top-K number of vertical and diagonal lines, and \textit{Block-Sparse} represents the Top-K number of blocks retained.}
    \begin{tabular}{l|c}
    \toprule
         Patterns & Search Space  \\
         \midrule
        A-shape & $\{(1024, 4096)\}$ \\
        Vertical-Slash & $\{(30, 2048), (100, 1800), (500, 1500), (3000, 200)\}$ \\
        Block-Sparse & $\{100\}$ \\
        \bottomrule
    \end{tabular}
    \label{tab:search_space}
\end{table}

\subsection{Single A100 Implementation Details}

The original PyTorch implementation\footnote{https://github.com/huggingface/transformers/blob/main/src/transformers/models/llama/modeling\_llama.py} of the LLaMA model causes an out-of-memory error on a single A100 (80G) when the prompt exceeds 50k tokens. To enable running 1M prompt inference on a single A100, we implemented the following optimizations:
\begin{enumerate}
    \item \textbf{Tensor Splitting}: We split the Attention by head and the MLP by sequence dimension. In long-context scenarios, where computation is the bottleneck, this splitting keeps GPU utilization at 100\%, and the overhead of splitting is negligible;
    \item \textbf{Reduction of Intermediate Variables}: We minimized intermediate variable allocation by removing the attention mask and implementing causal mask logic directly within the kernel;
    \item \textbf{Elimination of Unnecessary Computations}: In long-context scenarios, only the logits corresponding to the last token in the prompt phase are meaningful. Thus, we only retain the computation of the LM Head Linear layer for the last token.
\end{enumerate}

\subsection{Kernel Implementation}
\label{subsec:kernel}

\subsubsection{Block-Sparse Flash Attention}
Our \textit{Block-Sparse} kernel implementation is based on the Triton version of the FlashAttention kernel~\cite{triton-flash-attn}. With the selected block index as an additional input, each thread block loops through the top-K blocks in a row. As discussed in FlashAttention~\cite{dao2023flashattention}, the latency of the block-sparse FlashAttention kernel is linearly related to the number of blocks, and the speedup ratio (compared to the dense FlashAttention kernel) is approximately as,
\begin{equation}
    s_{p} = \frac{S}{2B\times k_b}
\end{equation}

\subsubsection{Vertical-Slash Attention}
The \textit{Vertical-Slash} attention includes two custom kernels: the \textit{Vertical-Slash} sparse index kernel and the \textit{Vertical-Slash} sparse FlashAttention kernel.

\begin{figure*}[htb]
  \vspace{-10pt}
  \centering
    \includegraphics[height=0.35\columnwidth]{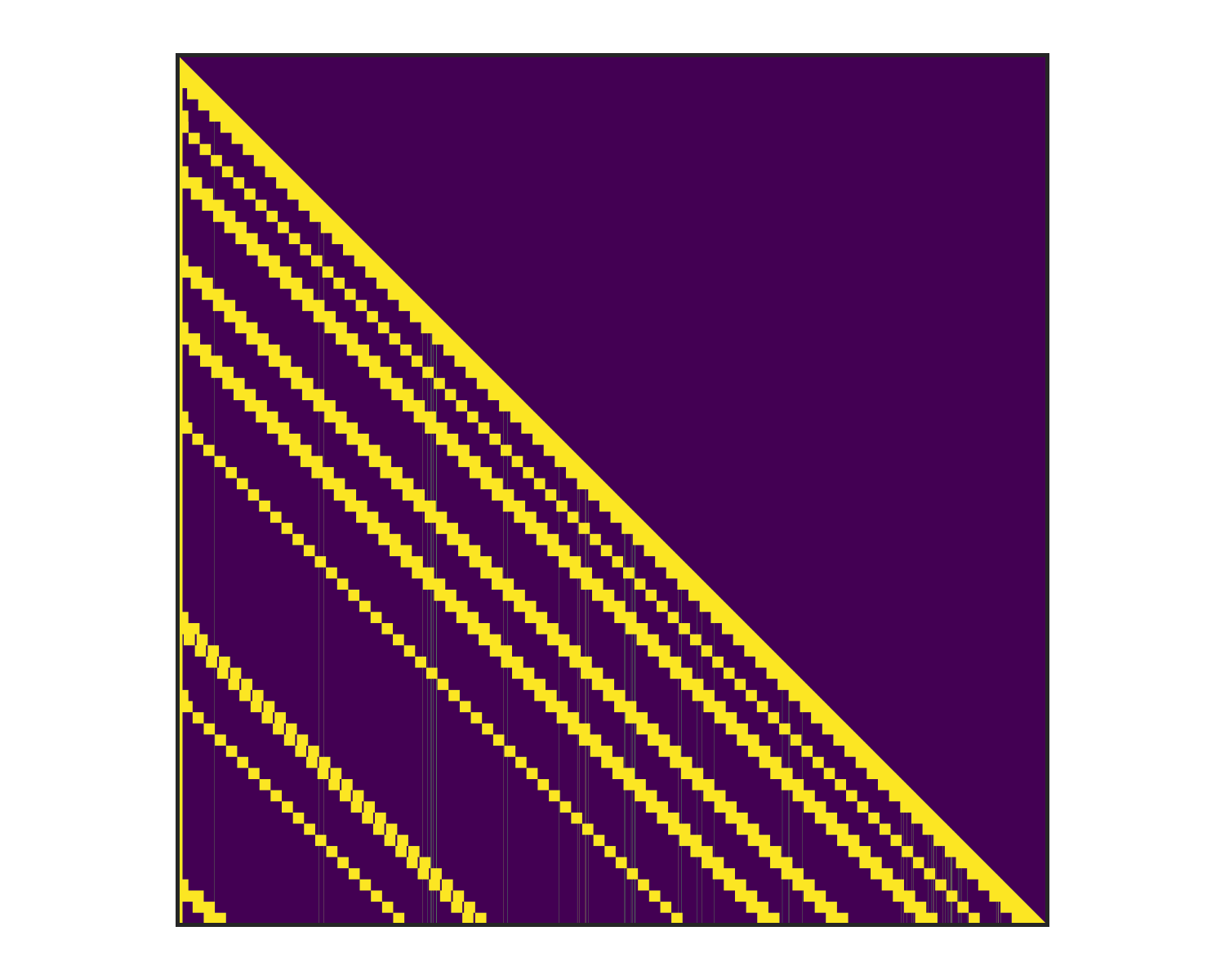}
  \caption{The dynamic sparse mask for the vertical-slash pattern using LLaMA-3-8B in the summarization task~\cite{zhang2024inftybench}. Yellow areas indicate the computed parts. Slash lines use $64\times64$ blocks, while vertical lines use $1\times64$ blocks.}
  \label{fig:vs_mask_pattern}
\end{figure*}

The \textit{Vertical-Slash} sparse index kernel in Algorithm~\ref{alg:vs_index} builds the index for each row of blocks. Since a slash line segment can be masked by a square block, our attention mask is a mix of blocks and columns, as shown in Fig.~\ref{fig:vs_mask_pattern}. We apply a point-range two-way merge algorithm where vertical indexes are treated as points and slash indexes are converted to ranges given the row index. The output consists of two parts: merged ranges and separate column indexes, where the ranges are represented by block indexes. The time complexity to build an index for a row is $O(k_v + k_s)$.

The \textit{Vertical-Slash} sparse FlashAttention kernel in Algorithm~\ref{alg:vs_attn} is a mix of the block-sparse attention kernel and the PIT~\cite{zheng2023pit} sparse attention kernel. PIT is a technology that loads sparse data into dense compute blocks via a Permutation Invariant Transformation. A thread block first loops through the block indexes as described in the previous section (block part) and then loops through the column indexes grouped by block size (PIT part). The latency of this hybrid kernel is linearly related to the total area of blocks and columns.

\section{Additional Experiment Results}
\label{sec:additional_experiment}

\subsection{Needle In A Haystack}
\label{subsec:needle}
\begin{wrapfigure}{r}{0.5\columnwidth}
    \vspace{-40pt}
    \centering
    \includegraphics[width=1\linewidth]{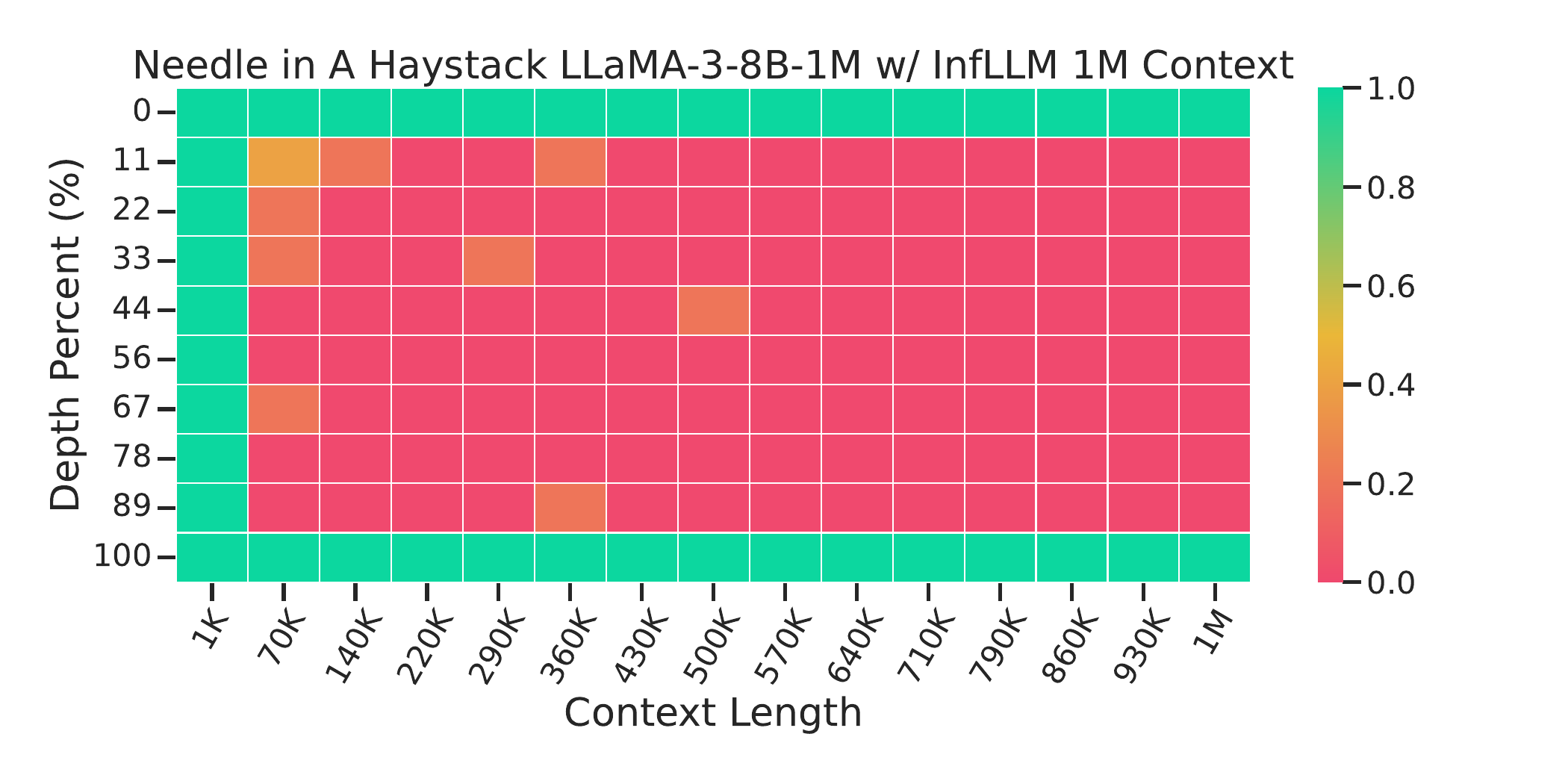}
    \caption{Results on Needle In A Haystack using InfLLM in LLaMA-3-8B-Instruct-1M.}
    \label{fig:needle_infllm}
    \vspace{-20pt}
\end{wrapfigure}

\begin{figure*}[htb]
  \centering
  \subfloat[GLM-4-9B-1M]{
    \label{sfig:needle_original_glm4}
    \includegraphics[width=0.49\columnwidth]{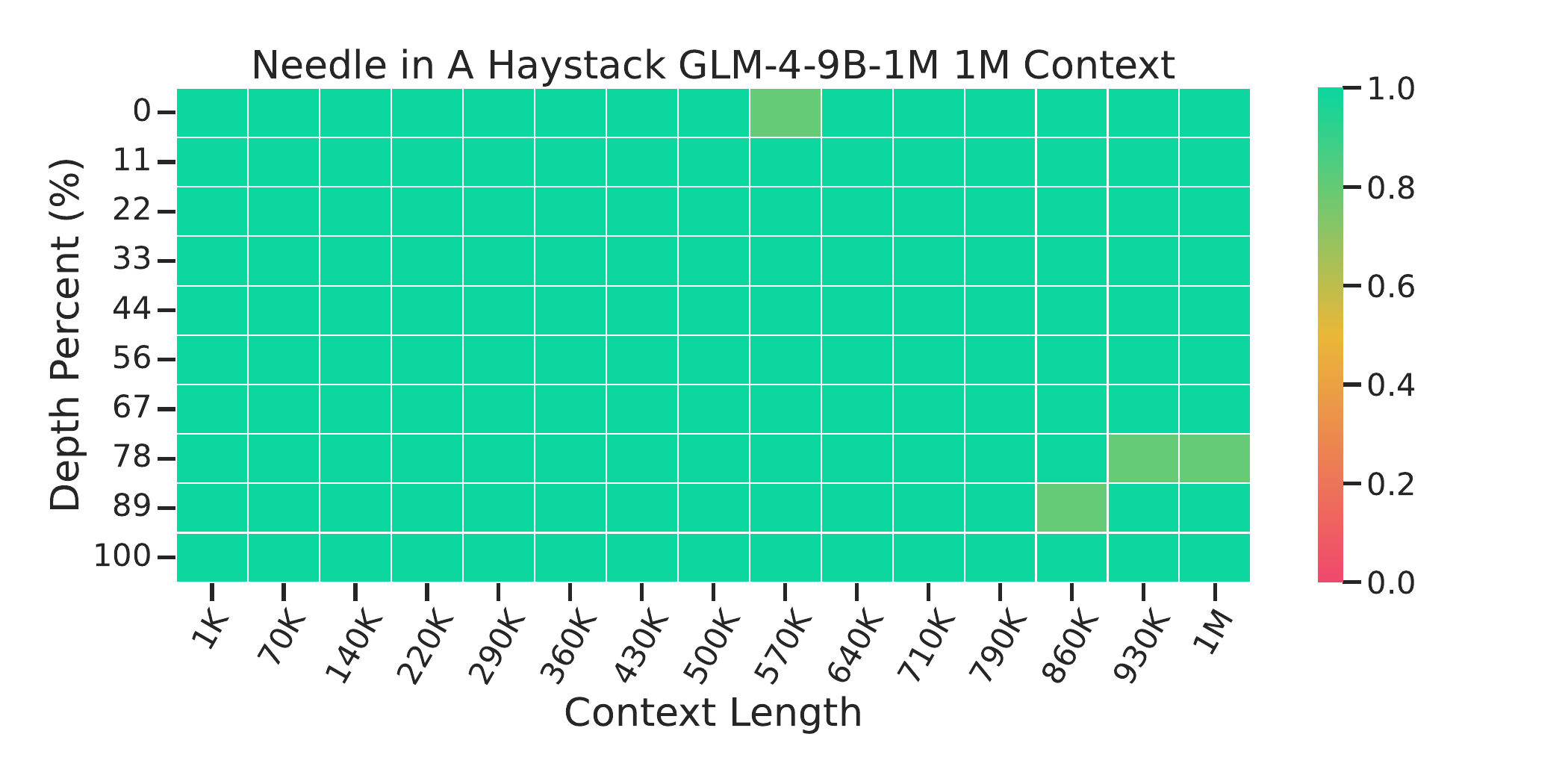}}
  \subfloat[GLM-4-9B-1M w/ MInference]{
    \label{sfig:needle_ours_glm4}
    \includegraphics[width=0.49\columnwidth]{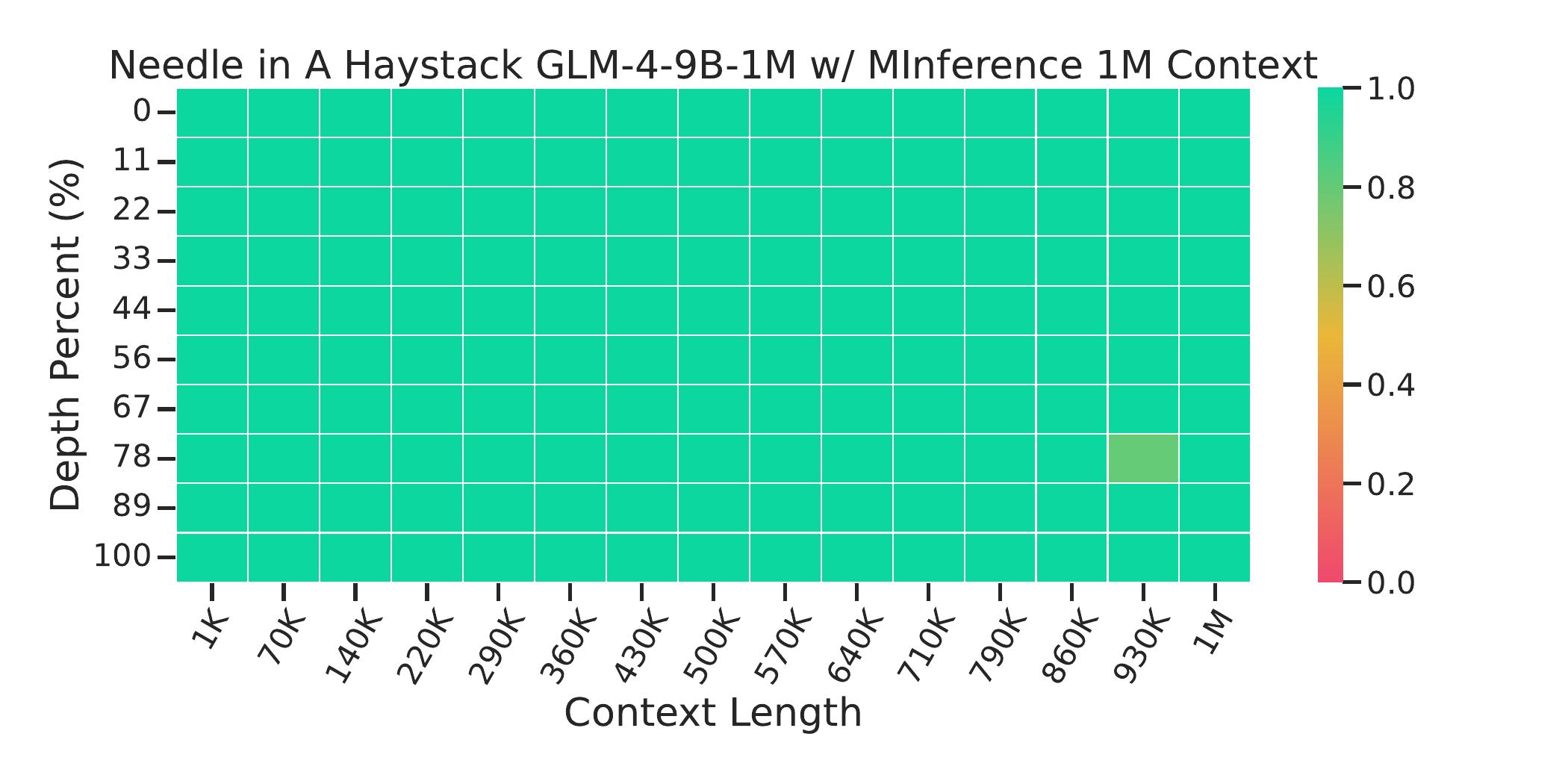}}\\
  \subfloat[Yi-9B-200K]{
    \label{sfig:needle_original_yi}
    \includegraphics[width=0.49\columnwidth]{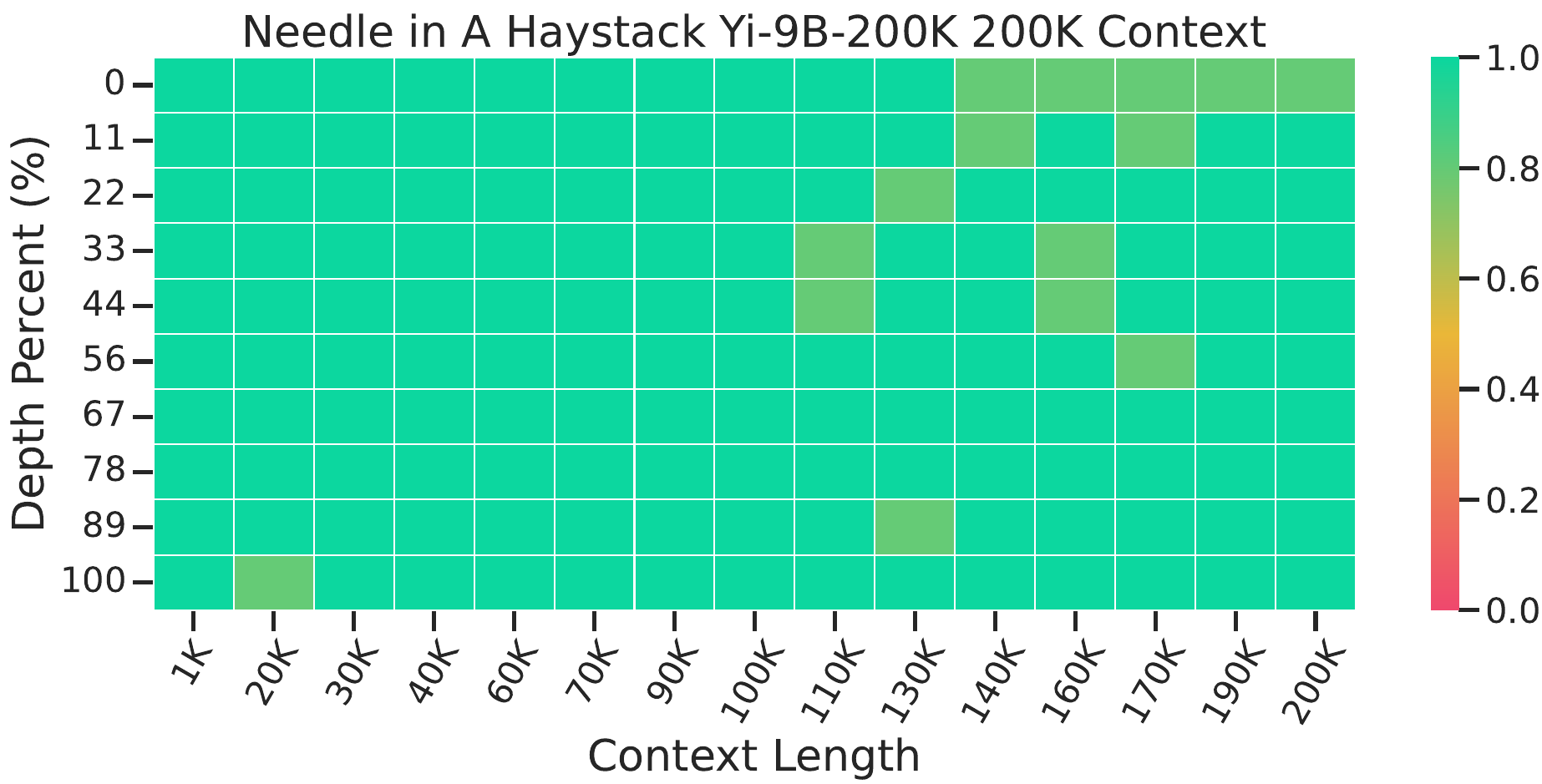}}
  \subfloat[Yi-9B-200K w/ MInference]{
    \label{sfig:needle_ours_yi}
    \includegraphics[width=0.49\columnwidth]{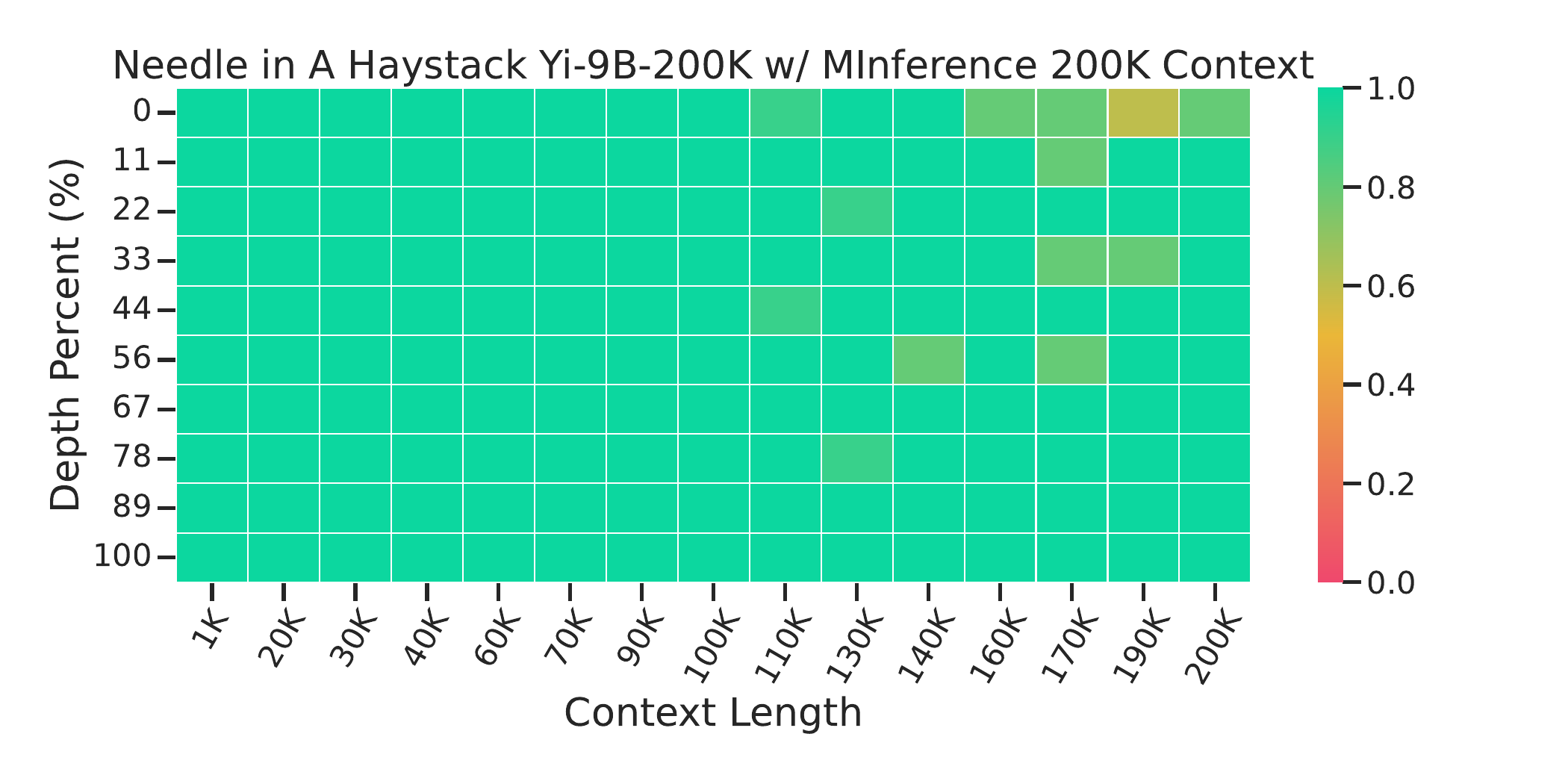}}\\
  \subfloat[Phi-3-Mini-128K]{
    \label{sfig:needle_original_phi3}
    \includegraphics[width=0.49\columnwidth]{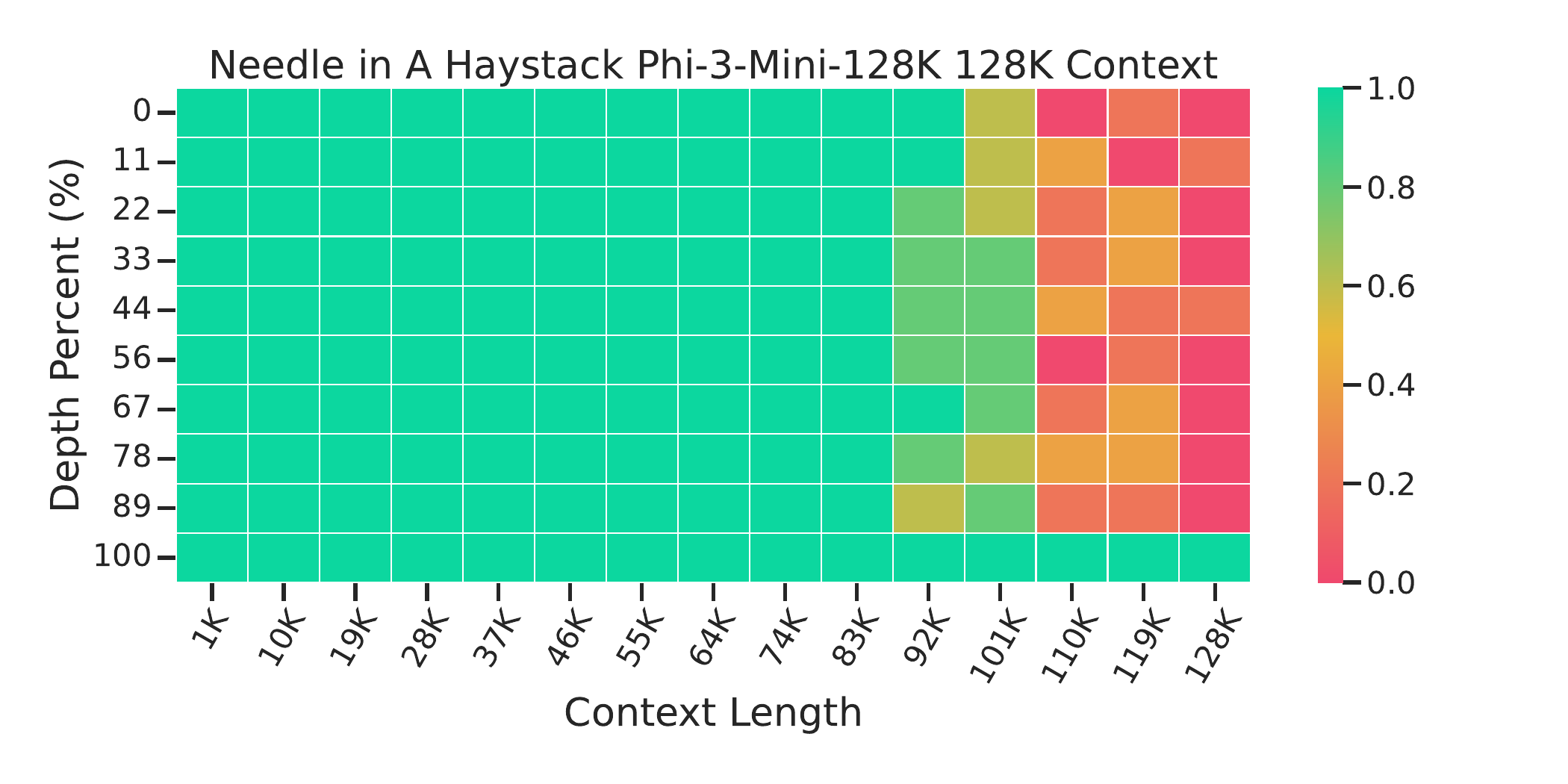}}
  \subfloat[Phi-3-Mini-128K w/ MInference]{
    \label{sfig:needle_ours_phi3}
    \includegraphics[width=0.49\columnwidth]{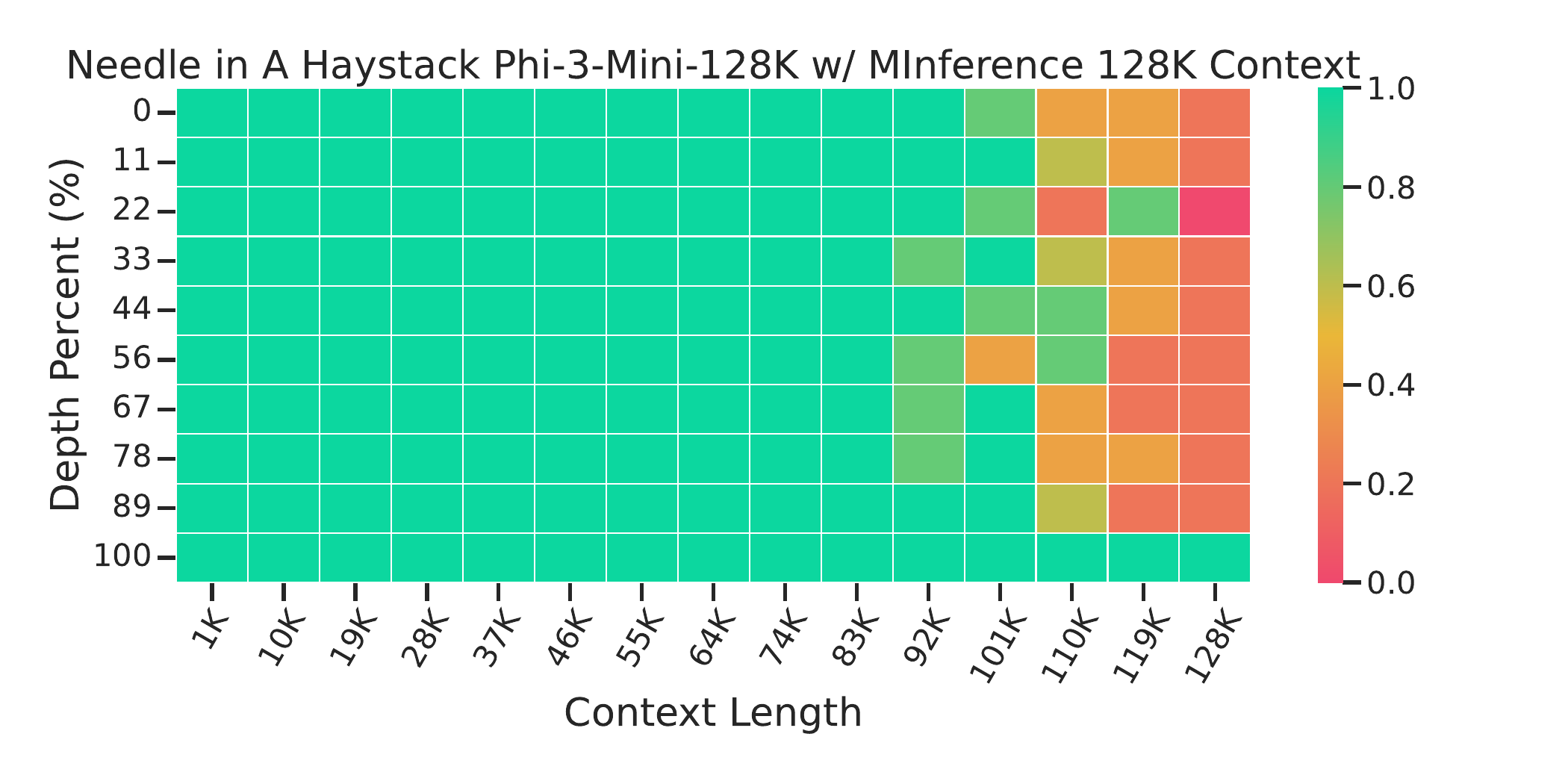}}\\
  \subfloat[Qwen2-7B-128K]{
    \label{sfig:needle_original_qwen2}
    \includegraphics[width=0.49\columnwidth]{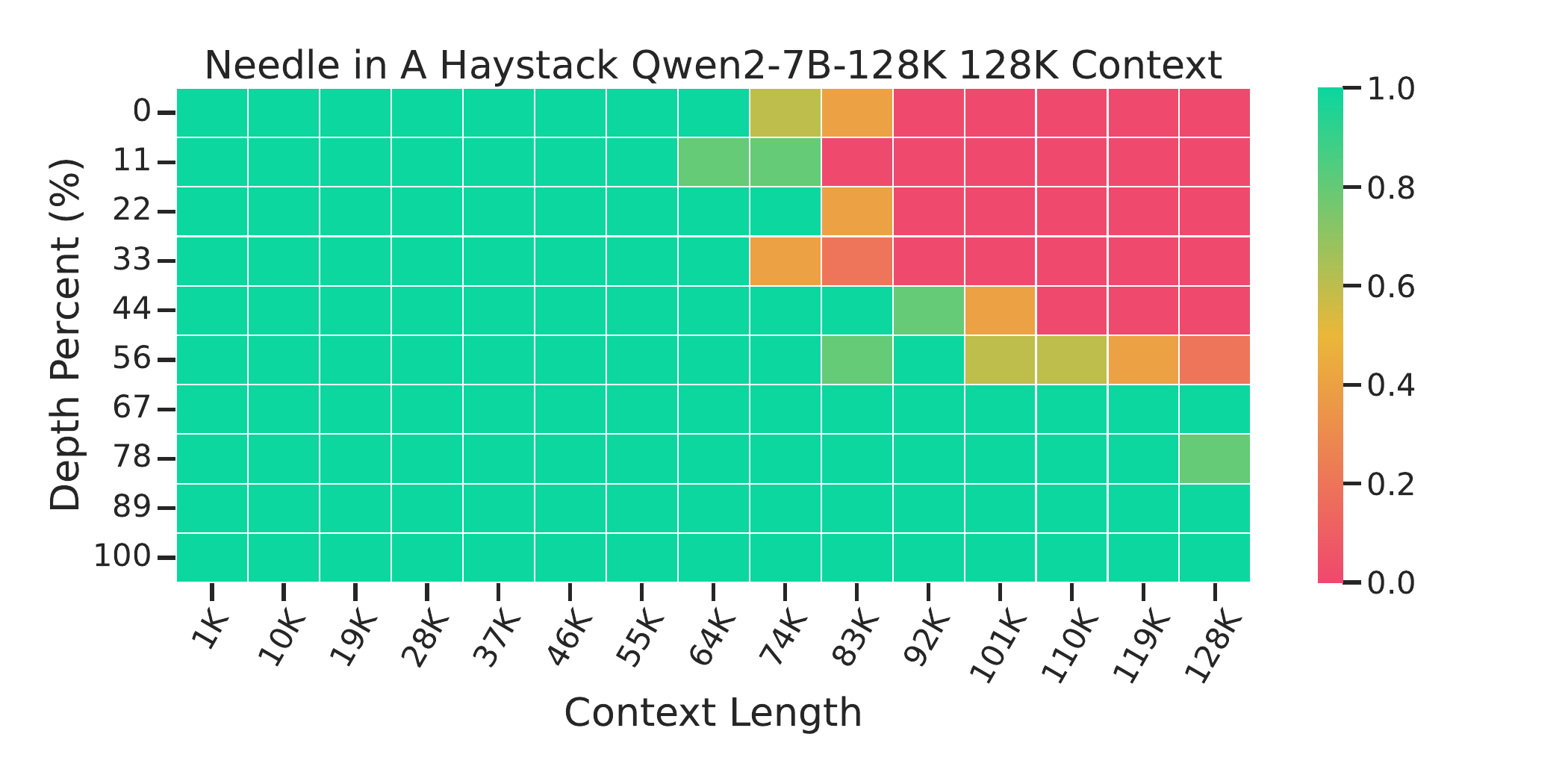}}
  \subfloat[Qwen2-7B-128K w/ MInference]{
    \label{sfig:needle_ours_qwen2}
    \includegraphics[width=0.49\columnwidth]{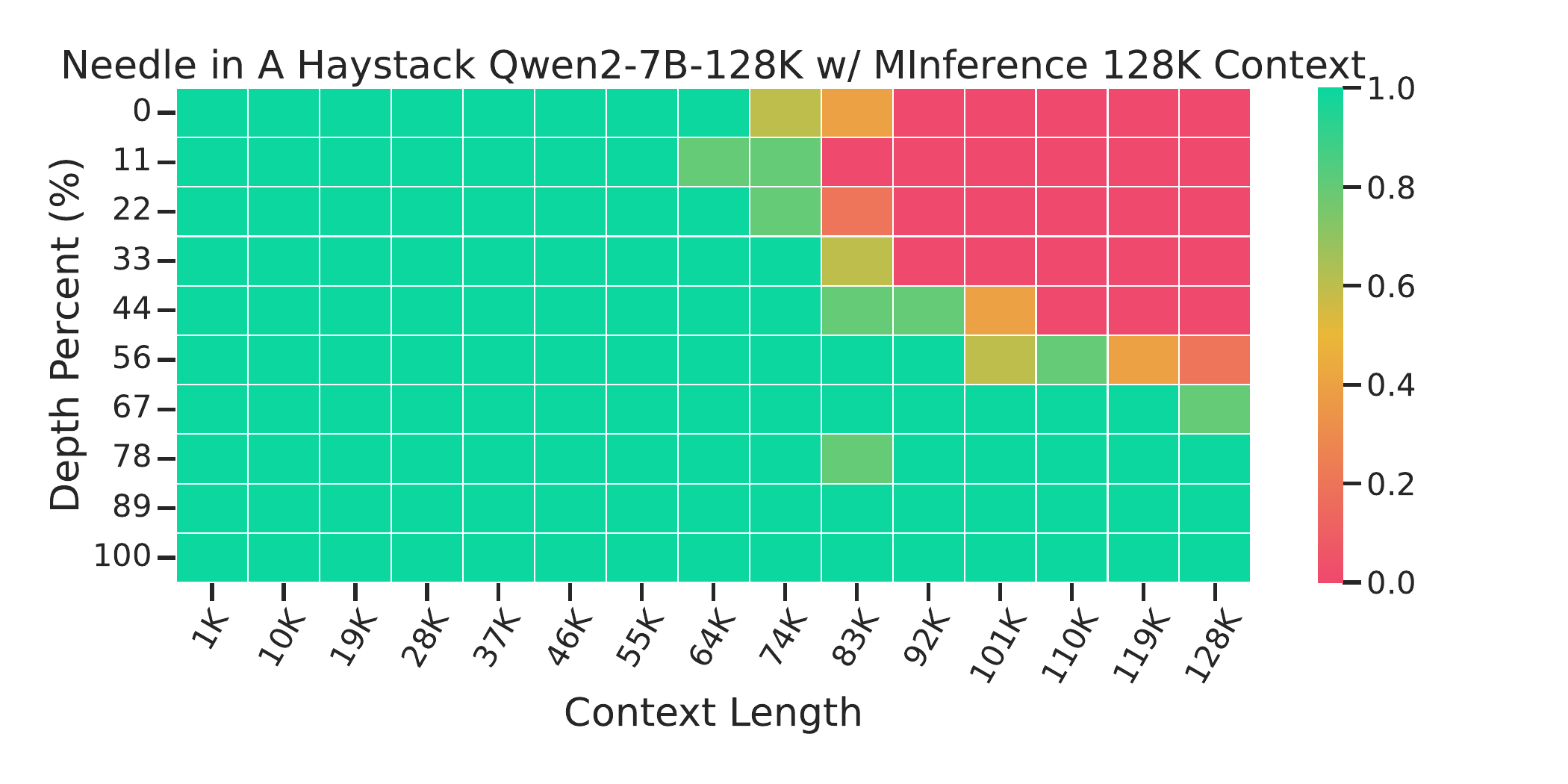}}\\
  \caption{Needle In A Haystack~\cite{kamradt2023needle} results using GLM-4-9B-1M~\cite{glm2024chatglm}, Yi-9B-200K~\cite{young2024yi}, Phi-3-Mini-128K~\cite{Abdin2024Phi3TR}, and Qwen2-7B-128K~\cite{qwen}.}
  \label{fig:needle_yi}
\end{figure*}

In addition to the Needle In A Haystack results for LLaMA-3-Instruct-1M shown in \S\ref{sec:experiments}, we also present the LLaMA-3-Instruct-1M using InfLLM results in Fig.~\ref{fig:needle_infllm}, and results for GLM-4-9B-1M, Yi-9B-200K, Phi-3-Mini-128K, and Qwen2-7B-128K, shown in Fig.~\ref{fig:needle_yi}. Compared to Full Attention, using MInference has minimal impact on the ability to understand semantic information across different context windows and needle depths. There is even a slight performance improvement around the 100k context length using Yi-9B-200K and Phi-3-Mini-128K.

\subsection{Latency Breakdown}

\begin{figure*}[htb]
  \centering
    \includegraphics[height=0.35\columnwidth]{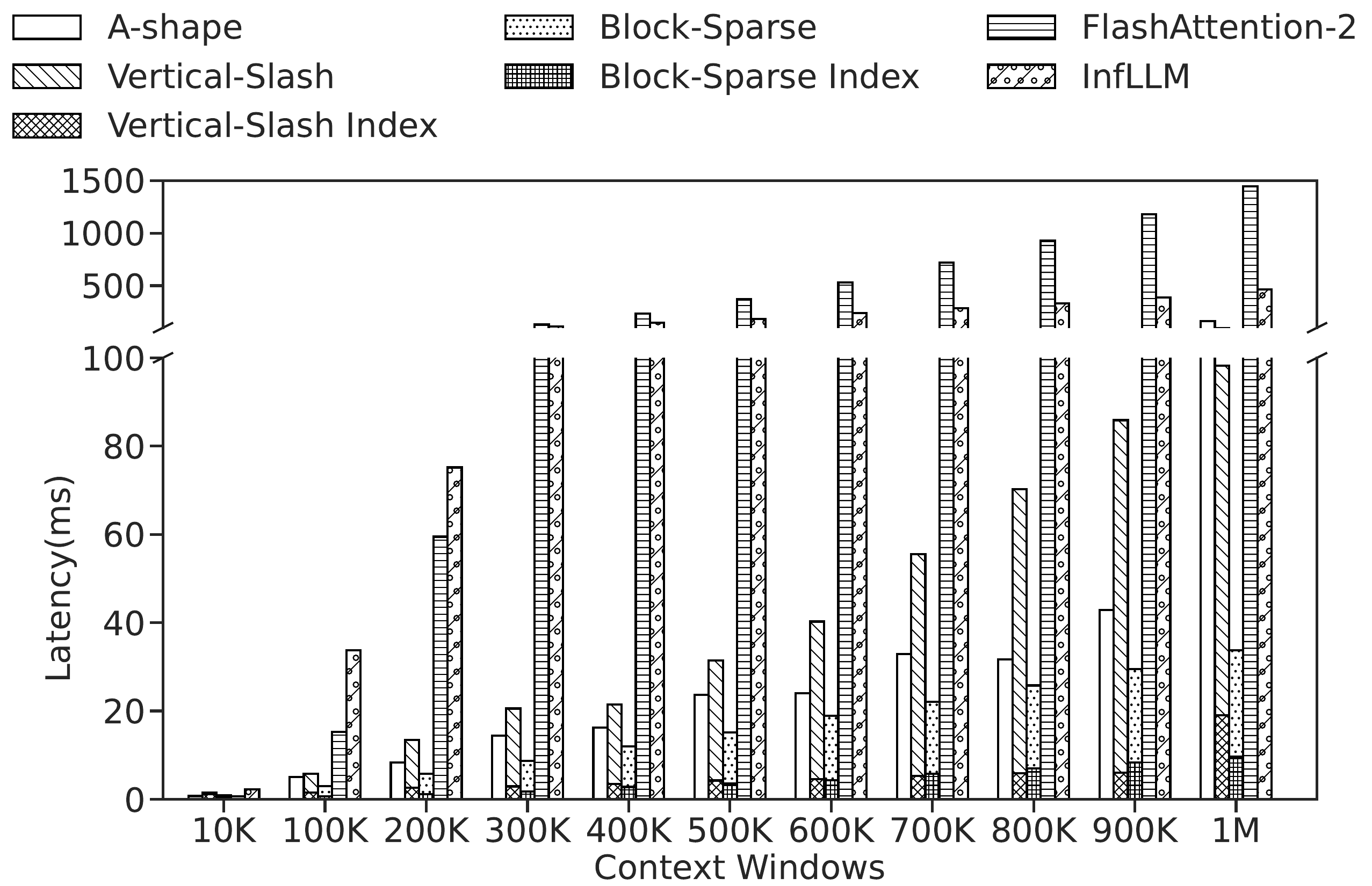}
  \caption{The latency breakdown of a single attention kernel for three patterns and FlashAttention~\cite{dao2023flashattention} across different context windows in a single A100, including the index time for dynamic sparse approximation and building dynamic sparsity. At 10k tokens, the latency of the four kernels is very close and all are less than 1ms. At 1M tokens, the latency for A-shape is 164ms.}
  \label{fig:latency_breakdown_detail}
\end{figure*}

Fig.~\ref{fig:latency_breakdown_detail} shows the micro-benchmark results of the three attention patterns proposed in this paper, as well as FlashAttention. It can be seen that Vertical-Slash is the slowest among the three patterns, but it still achieves a 13x speedup compared to FlashAttention under 1M context windows. A-shape is slightly faster than Vertical-Slash, but at 1M, A-shape is 50\% slower than Vertical-Slash. Block-Sparse is the fastest, achieving a 30x speedup over FlashAttention under 1M context windows. The estimation and index-building time for the dynamic sparse pattern accounts for approximately 5\%-15\% and 25\% of the total time for Vertical-Slash and Block-Sparse patterns, respectively. The index-building overhead is higher for Block-Sparse mainly due to the time-consuming MeanPooling and block-level matmul computations. Additionally, the memory overhead for sparse indexing is relatively small, remaining within 160MB for a LLaMA-3-8B model in 1M context.

\subsection{Additional Ablation Study}
\begin{table}[htb]
    \small
    \centering
    \setlength{\tabcolsep}{0.5mm}
    \vspace{-2ex}
    \caption{Performance of different ablation methods using LLaMA-3-8B-Instruct-262K on InfiniteBench~\cite{zhang2024inftybench}.
    It is important to note that due to kernel limitations, we must retain at least one vertical and one slash. Therefore, "ours w/ only vertical" retains the top-1 slash, and "ours w/ only slash" retains the top-1 vertical.
    }
    \resizebox{\columnwidth}{!}{
    \begin{tabular}{l|cccccccccc|c}
    \toprule
        Methods &  En.Sum & En.QA & En.MC & En.Dia & Zh.QA & Code.Debug & Math.Find & Retr.PassKey & Retr.Num & Retr.KV &  Avg. \\
    \midrule
    
    Ours & 20.5 & 12.9 & 65.9 & 7.5 & 12.5 & 22.3 & 33.1 & 100.0 & 100.0 & 12.8 & 38.8 \\
    Ours w/ only vertical & 13.7 & 6.2 & 30.1 & 2.0 & 6.5 & 7.9 & 1.7 & 65.4 & 52.7 & 0.0 & 18.6 \\
    Ours w/ only slash & 18.4 & 11.5 & 60.1 & 3.0 & 11.4 & 22.1 & 28.4 & 100.0 & 100.0 & 4.2 & 35.9 \\
    \bottomrule
    \end{tabular}
    }
    \label{tab:additional_ablation}
\end{table}

To further analyze the role of dynamic vertical and slash lines in the \textit{Vertical-Slash} pattern for sparse computation, we introduce a new set of ablation studies as follows:
1) Ours w/ only vertical, which only uses vertical lines and the top-1 slash line in \textit{Vertical-Slash} pattern.
2) Ours w/ only slash, which only uses slash lines and the top-1 vertical line in \textit{Vertical-Slash} pattern.
The corresponding top-K quantities are set after converting based on FLOPs in kernel.

As shown in Table~\ref{tab:additional_ablation}, using only vertical lines results in a significant performance drop, especially in retrieval tasks, where performance is similar to only using block-sparse. In contrast, using only slash lines retains most of the performance, but in highly dynamic tasks such as KV retrieval, performance further decreases, with an average performance drop of 2.9\% compared to Ours.

\section{Pattern Distribution}
\label{sec:pattern}

\begin{figure*}[htb]
  \centering
  \subfloat[LLaMA-3-8B-Instruct-262K/1M]{
    \label{sfig:pattern_llama}
    \includegraphics[height=0.27\columnwidth]{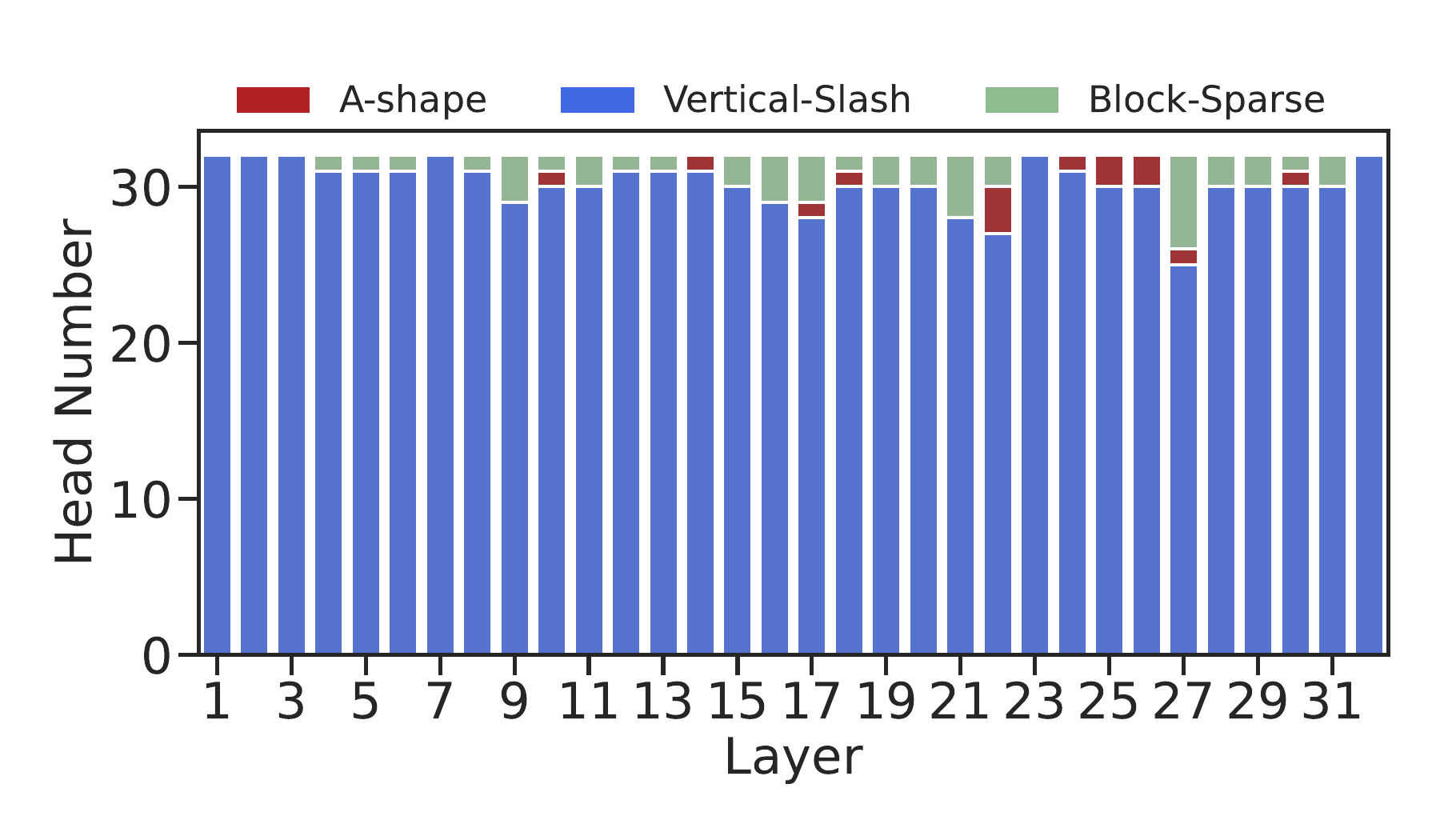}}
  \subfloat[Yi-9B-200K]{
    \label{sfig:pattern_yi}
    \includegraphics[height=0.27\columnwidth]{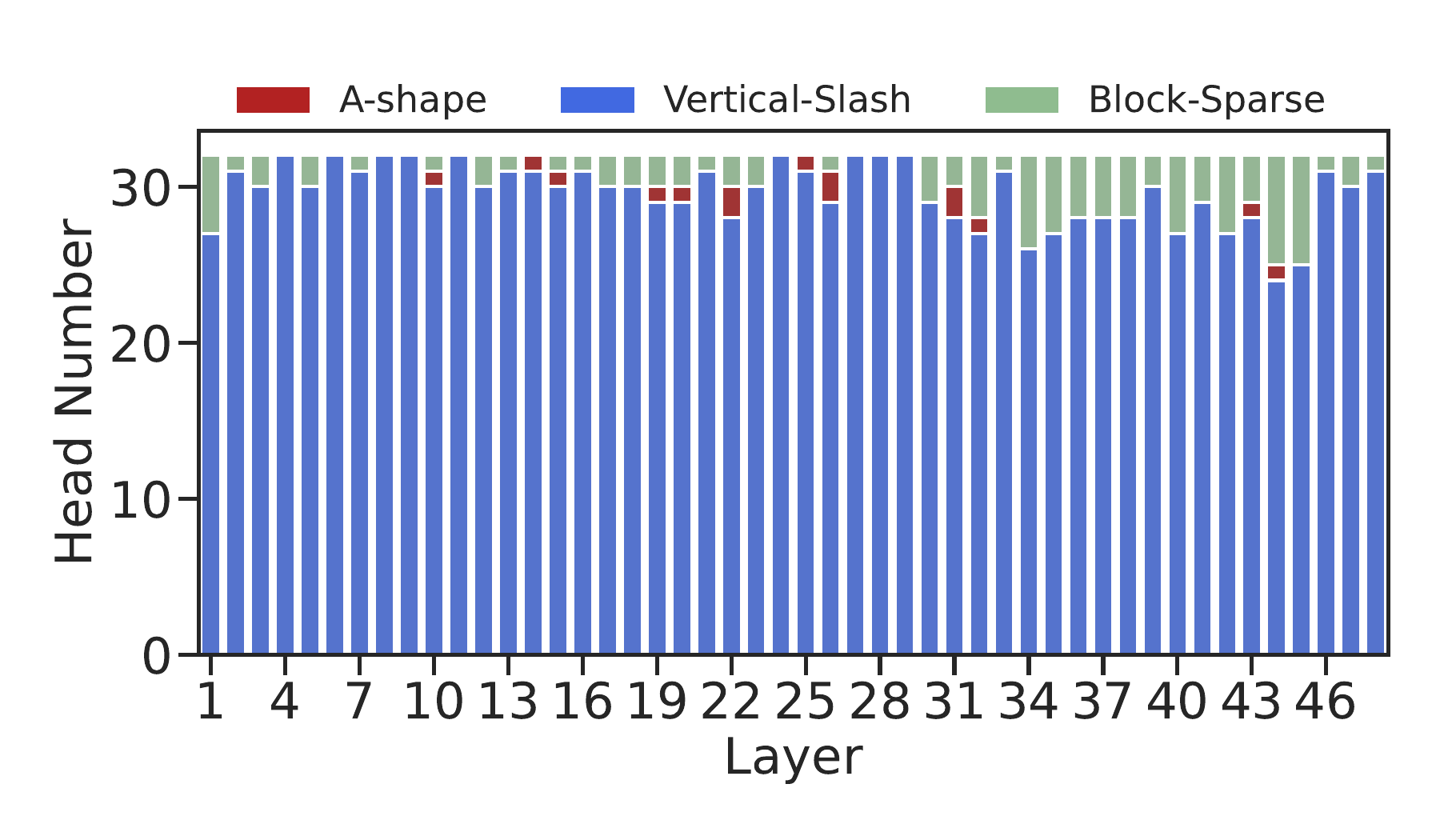}}
  \caption{Distribution of three sparse head patterns in different models. We use the same optimal sparse pattern configuration for both LLaMA-3-8B-Instruct-262K and LLaMA-3-8B-Instruct-1M.}
  \label{fig:pattern_distribution}
\end{figure*}

Fig.~\ref{fig:pattern_distribution} shows the distribution of the optimal head configuration obtained through our search. Firstly, most of the patterns are the \textit{Vertical-Slash} pattern (>90\%). However, according to the ablation study, using only the \textit{Vertical-Slash} pattern significantly impacts performance in highly dynamic tasks like KV retrieval. Secondly, the \textit{Block-Sparse} pattern is primarily distributed in several intermediate to later layers, while the \textit{A-shape} pattern is found in the middle layers. Although the optimal patterns vary slightly across different models, they generally align with these observations.

Additionally, we used the same configuration for two versions of LLaMA in our experiments, and the results show that the 1M model also performs very well, with nearly perfect results in the Needle In A Haystack task. This demonstrates the generalizability of the optimal sparse pattern.

\section{Sparsity in Kernel Distribution}
\label{sec:sparsity}
\begin{figure*}[!h]
  \centering
    \includegraphics[height=0.38\columnwidth]{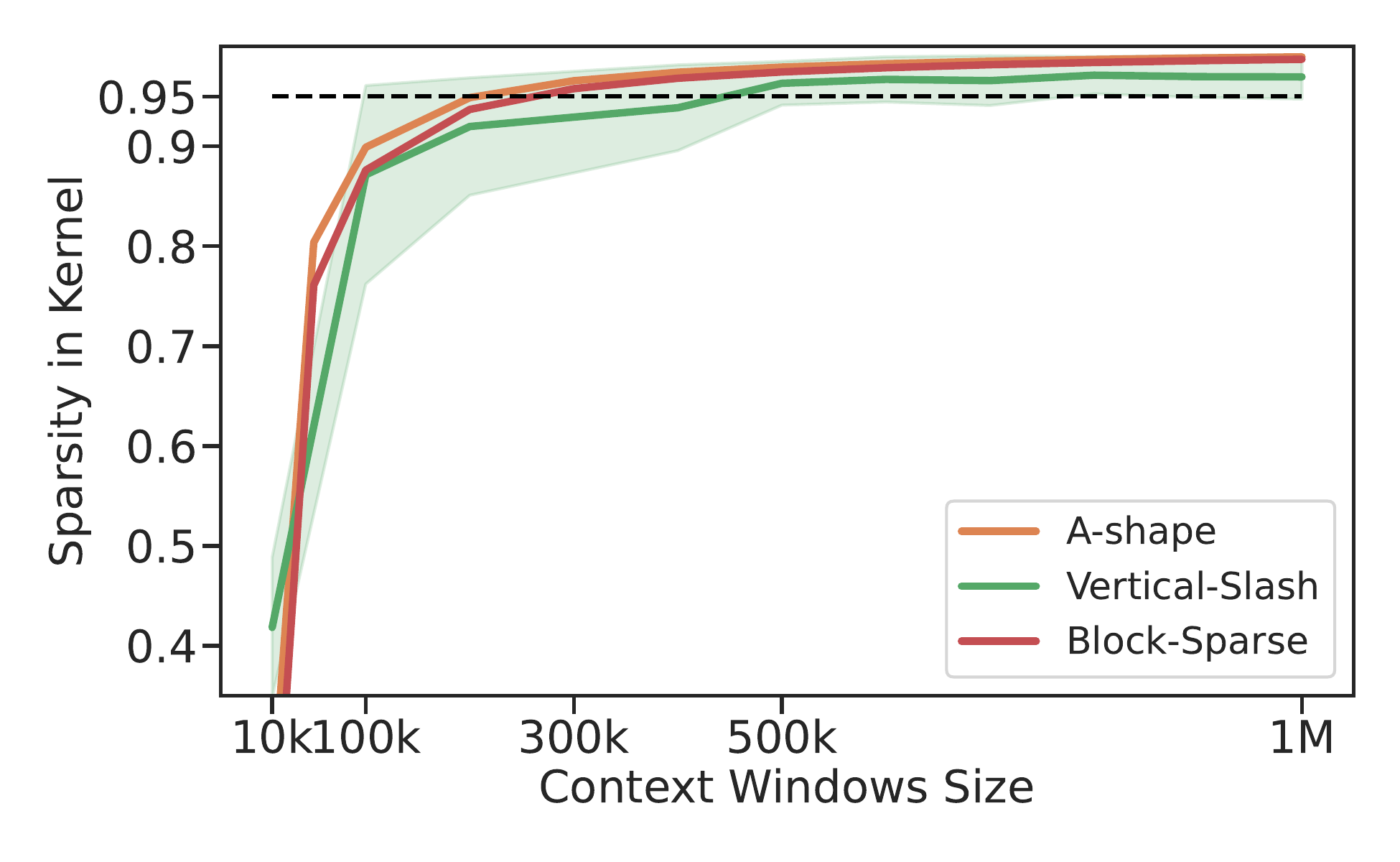}
  \caption{The distribution of sparsity in the kernel across different context windows refers to the proportion of the kernel that is actually computed after block coverage, compared to the sparsity rate when using FlashAttention with a causal mask.}
  \label{fig:sparsity_in_kernel}
\end{figure*}

As shown in Fig.~\ref{fig:sparsity_in_kernel}, the sparsity distribution of the three patterns during the actual kernel computation process is displayed. It can be seen that when the context windows exceed 200k, the actual sparsity of all three patterns surpasses 90\%. Even considering a 20\% index-building overhead, this ensures that the kernel achieves a speedup of over 8$\times$. Furthermore, when the context windows exceed 500k, the sparsity relative to FlashAttention exceeds 95\%, with a theoretical speedup of over 15$\times$.

\section{Does This Dynamic Sparse Attention Pattern Exist Only in Auto-Regressive LLMs or RoPE-Based LLMs?}
\label{sec:encoder_multi_modal}

\begin{figure*}[htb]
  \centering
    \includegraphics[height=0.35\columnwidth]{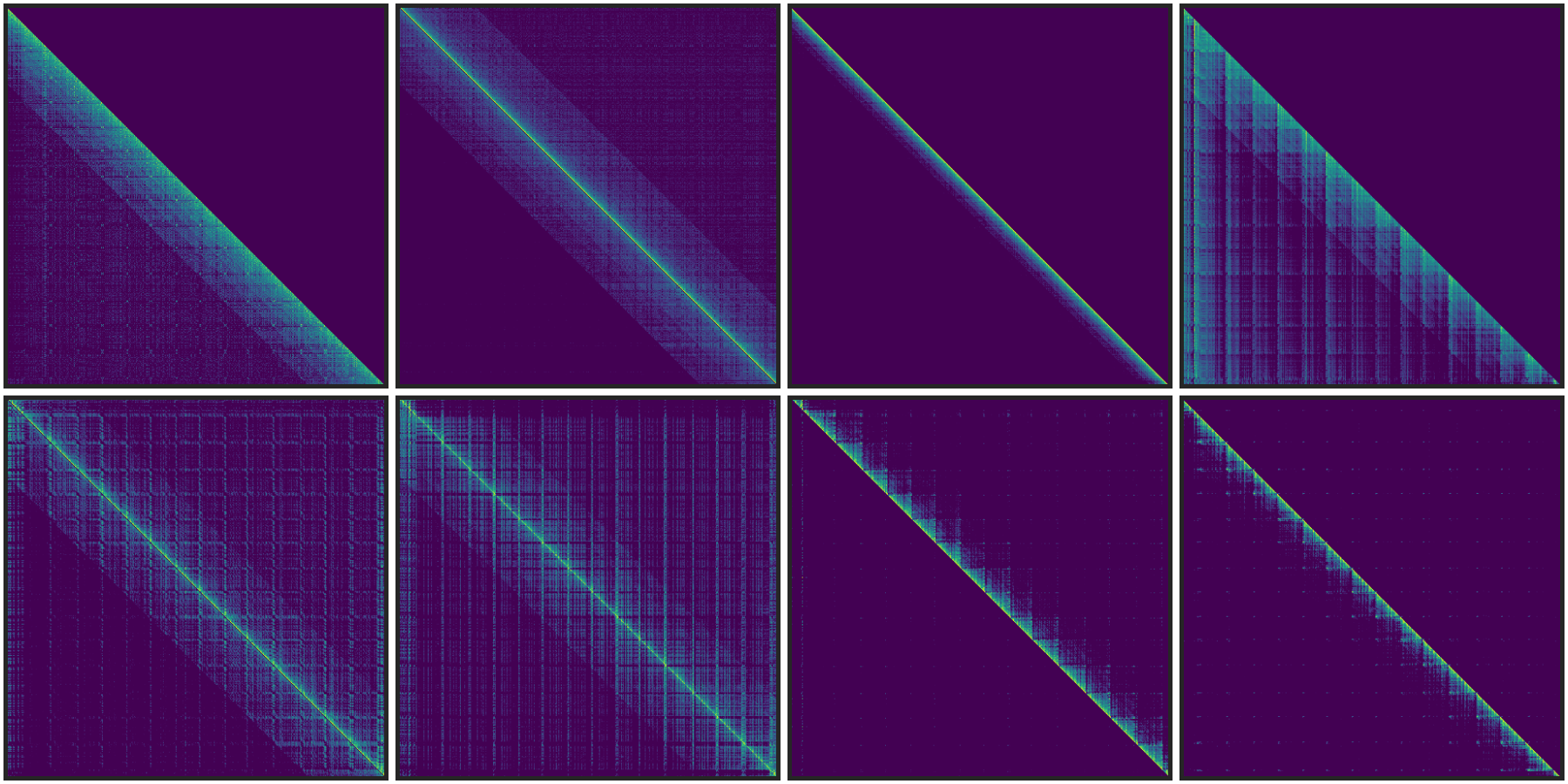}
  \caption{The sparse pattern in T5-style Encoder Attention using Flan-UL2~\cite{tay2023ul} on the Summarization dataset~\cite{zhang2024inftybench}.}
  \label{fig:sparsity_t5}
\end{figure*}

Similar vertical and slash line sparse patterns have been discovered in BERT~\cite{shi2021sparsebert} and multi-modal LLMs~\cite{wan2024lookmlookonceoptimizationkv}. Additionally, as shown in Fig.~\ref{fig:sparsity_t5}, we analyzed the distribution of attention patterns in T5 across different heads. It is evident that there are vertical and slash sparse patterns even in bidirectional attention.

Recent studies~\cite{wan2024lookmlookonceoptimizationkv} have analyzed sparse attention patterns in multi-modal LLMs, revealing the presence of vertical and slash patterns in models like LLaVA~\cite{liu2024visual} and InternVL~\cite{chen2024internvl}. Using MInference for pre-filling stage inference acceleration holds great promise.

\section{Case Study}

Table~\ref{tab:case_summary} presents a comparison of the generation performance for various methods on the EN.SUM task (200K input length) from InfiniteBench based on the LLaMA-3-8B-262K model. The original summary provides a comprehensive and coherent narrative, detailing the Bronwyn family's trip to the Kindergarten and touching on themes such as nostalgia, loss, and the passage of time. StreamingLLM's summary, although looks coherent, introduces elements that are not present in the original story, leading to serious factual errors. For example, it mentions a boat trip to a school for boys and specific details like fishermen, sandwiches, and a spot where men were drowned. These details deviate from the original story, which is about the Bronwyn family preparing for a trip to the Kindergarten. In addition, the summaries generated by StreamingLLM with dilated and strided techniques are largely incoherent, consisting primarily of repetitive and nonsensical characters, indicating a failure to produce meaningful content. In stark contrast, the summary generated by our proposed method offers a detailed and coherent narrative, comparable to the original, with a clear depiction of the story's main events and themes. This includes the preparation of the Bronwyn family for their trip, the characterization of family members and guests, and the exploration of deeper themes such as love, marriage, and the search for meaning. The results demonstrate the superiority of our proposed method in generating high-quality, human-like summaries over the baseline methods.

Table~\ref{tab:case_kv} compares the performance of various methods on the Retrieve.KV task (200K input length) using the LLaMA-3-8B-262K model. The original method demonstrates perfect retrieval, correctly predicting the exact strings of the ground truth for both examples. StreamingLLM, again, generates predictions that looks coherent and real, but factually incorrect. In addition, StreamingLLM with dilated and strided techniques, and our method with a static pattern, fail significantly, producing outputs that are either repetitive sequences of characters or nonsensical strings, indicating their inability to accurately retrieve the required key-value pairs. Our method, however, performs on par with the original, accurately retrieving and predicting the exact key-value pairs for both examples. This demonstrates the superior capability of our method in handling KV retrieval tasks, providing precise and reliable outputs consistent with the ground truth. The results highlight our method's effectiveness and robustness compared to the baselines, making it a reliable choice for such tasks.

\restylefloat{algorithm}

\begin{figure}[htb]

\centering
\begin{minipage}[t]{0.5\textwidth}
\vspace{0pt}
\centering
\begin{algorithm}[H]
\captionsetup[algorithm]{singlelinecheck=off}
\caption{Vertical-Slash Index}
\label{alg:vs_index}
\begin{algorithmic}
    \STATE {\bfseries Input:} vertical indexes $\boldsymbol{i}_v \in \mathbb{N}^{k_v}$, slash indexes $\boldsymbol{i}_s \in \mathbb{N}^{k_s}$

    \LineComment{Sort vertical and slash indexes}

    \STATE $\boldsymbol{i}_v \gets \mathrm{IncrementalSort}\left(\boldsymbol{i}_v\right)$
    \STATE $\boldsymbol{i}_s \gets \mathrm{DescendingSort}\left(\boldsymbol{i}_s\right)$

    \LineComment{Calculate block number (block\_size $B$)}
    \STATE $N \gets \lceil\frac{S}{B}\rceil$

    \LineComment{Initialize outputs}
    \STATE block count $\boldsymbol{c}_{\text{blk}} \in \mathbb{N}^{N}$, block index $\boldsymbol{i}_{\text{blk}}\in \mathbb{N}^{N \times k_s}$, column count $\boldsymbol{c}_{\text{col}} \mathbb{N}^{N}$, column index $\boldsymbol{i}_{\text{col}} \in \mathbb{N}^{N \times k_v}$

    \LineComment{Parallelized in GPU}
    \FOR{$i \gets 1$ to $N$}
        \STATE $j_v \gets 1$

        \LineComment{Find the first slash line that crosses the row}
        \STATE $j_s \gets \text{biset\_left}(\boldsymbol{i}_s, i \times B)$

        \LineComment{Define the range by slash index}
        \STATE $r_{\text{start}} \gets (i - 1) \times B - \boldsymbol{i}_s^{j_s}$
        \STATE $r_{\text{end}} \gets i \times B - \boldsymbol{i}_s^{j_s}$

        \LineComment{Merge points (vertical indexes) and ranges (slash indexes)}
        \WHILE{$s_v \leq k_s$}
            \IF{$j_v \leq k_v$ \AND $\boldsymbol{i}_v^{j_v} < r_{\text{end}}$}
                \STATE
                \LineComment{Record the point if not in the range}
                \IF{$\boldsymbol{i}_v^{j_v} < r_{\text{start}}$}
                    \STATE $\boldsymbol{c}_{\text{col}}^i \gets \boldsymbol{c}_{\text{col}}^i + 1$, $\boldsymbol{i}_{\text{col}}^{i,\boldsymbol{c}_{\text{col}}^i} \gets \boldsymbol{i}_v^{j_v}$
                \ENDIF
                \STATE $j_v \gets j_v + 1$
            \ELSE
                \STATE $s_v \gets s_v + 1$

                \LineComment{Update the range}
                \IF{$(i - 1) \times B - \boldsymbol{i}_s^{j_s} > r_{\text{end}}$}

                    \STATE
                    \LineComment{Record the last range}
                    \STATE $s \gets r_{\text{start}}$
                    \WHILE{$s < r_{\text{end}}$}
                        \STATE $\boldsymbol{c}_{\text{blk}}^i \gets \boldsymbol{c}_{\text{blk}}^i + 1$
                        \STATE $\boldsymbol{i}_{\text{blk}}^{i,\boldsymbol{c}_{\text{blk}}^i} \gets s$
                        \STATE $s \gets s + B$
                    \ENDWHILE

                    \LineComment{Calculate the new range}
                    \STATE $r_{\text{start}} \gets (i - 1) \times B - \boldsymbol{i}_s^{j_s}$
                    \STATE $r_{\text{end}} \gets i \times B - \boldsymbol{i}_s^{j_s}$
                \ELSE
                    \STATE
                    \LineComment{Extend the range}
                    \STATE $r_{\text{end}} \gets r_{\text{end}} + B$
                \ENDIF
            \ENDIF
        \ENDWHILE

        \LineComment{Record the last range}
        \STATE $s \gets r_{\text{start}}$
        \WHILE{$s < r_{\text{end}}$}
            \STATE $\boldsymbol{c}_{\text{blk}}^i \gets \boldsymbol{c}_{\text{blk}}^i + 1$
            \STATE $\boldsymbol{i}_{\text{blk}}^{i,c_{blk}^i} \gets s$, $s \gets s + B$
        \ENDWHILE
    \ENDFOR

    \STATE $\mathrm{return}\,\,\,\boldsymbol{c}_{\text{blk}},\boldsymbol{i}_{\text{blk}},\boldsymbol{c}_{\text{col}},\boldsymbol{i}_{\text{col}}$
\end{algorithmic}
\end{algorithm}
\end{minipage}
\hfill
\begin{minipage}[t]{0.48\textwidth}
\vspace{0pt}
\centering
\begin{algorithm}[H]
\captionsetup[algorithm]{singlelinecheck=off}
\caption{Vertical-Slash Flash Attention}
\label{alg:vs_attn}
\begin{algorithmic}
    \STATE {\bfseries Input:} $\boldsymbol{Q}, \boldsymbol{K}, \boldsymbol{V} \in \mathbb{R}^{S \times d_h}$, block count $\boldsymbol{c}_{\text{blk}} \in \mathbb{N}^{N}$, block index $\boldsymbol{i}_{\text{blk}} \in \mathbb{N}^{N \times k_s}$, column count $\boldsymbol{c}_{\text{col}} \in \mathbb{N}^{N}$, column index $\boldsymbol{i}_{\text{col}} \in \mathbb{N}^{N \times k_v}$

    \STATE Scale $\tau \gets \sqrt{\frac1{d_h}}$
    \STATE Initialize $\boldsymbol{O} \gets (0)^{S \times d_h} \in \mathbb{R}^{S \times d_h}$

    \LineComment{Parallelized in GPU}
    \FOR{$i \gets 1$ to $N$}
        \STATE Load $\boldsymbol{Q}_{\text{chip}} 
        \gets \boldsymbol{Q}^{i \times B:(i + 1)\times B} \in \mathbb{R}^{B \times d_h}$
        \STATE Initialize $\boldsymbol{O}_{\text{chip}} \gets (0)^{B \times d_h} \in \mathbb{R}^{B \times d_h}$
        \STATE Initialize $\boldsymbol{m} \gets (-\inf)^{B} \in \mathbb{R}^{B}$
        \STATE Initialize $\boldsymbol{l} \gets (0)^{B} \in \mathbb{R}^{B}$

        \LineComment{Loop through block indexes: block sparse flash attention}
        \FOR{$j \gets 1$ to $\boldsymbol{c}_{\text{blk}}^i$}
            \STATE Block start $s \gets \boldsymbol{i}_{\text{blk}}^{i,j}$
            \STATE Load $\boldsymbol{K}_{\text{chip}}
        \gets \boldsymbol{K}^{s:s+B} \in \mathbb{R}^{B \times d_h}$
            \STATE Load $\boldsymbol{V}_{\text{chip}}
        \gets \boldsymbol{V}^{s:s+B} \in \mathbb{R}^{B \times d_h}$
            \STATE $\boldsymbol{S} \gets \tau\boldsymbol{Q}_{\text{chip}}\boldsymbol{K}_{\text{chip}}^T$
            \STATE $\boldsymbol{S} \gets \mathrm{mask}(\boldsymbol{S})$
            \STATE $\boldsymbol{m}^i_{new} \gets \mathrm{max}(\boldsymbol{m}^i, \mathrm{rowmax}(\boldsymbol{S})) \in \mathbb{R}^{B}$
            \STATE $\boldsymbol{S} \gets \boldsymbol{S} - \boldsymbol{m}^i_{new}$
            \STATE $\boldsymbol{P} \gets \mathrm{exp}(\boldsymbol{S})$
            \STATE $\boldsymbol{l}^i_{new} \gets \mathrm{rowsum}(\boldsymbol{S}))$
            \STATE $\boldsymbol{\alpha} \gets \mathrm{exp}(\boldsymbol{m}^i - \boldsymbol{m}^i_{new})$
            \STATE $\boldsymbol{l}^i \gets \boldsymbol{\alpha}\boldsymbol{l}^i + \boldsymbol{l}^i_{new}$
            \STATE $\boldsymbol{O}_{\text{chip}} \gets \boldsymbol{\alpha}\boldsymbol{O}_{\text{chip}} + \boldsymbol{P}\boldsymbol{V}_{\text{chip}}$
        \ENDFOR

        \LineComment{Loop through column indexes : PIT sparse flash attention}
        \STATE $j \gets 0$
        \WHILE{$j < \boldsymbol{c}_{\text{col}}^j$}
            \STATE $\boldsymbol{cols} \gets \boldsymbol{i}_{\text{col}}^{i,j:j+B} \in \mathbb{N}^{B}$
            \STATE Load $\boldsymbol{K}_{\text{chip}}
        \gets \boldsymbol{K}^{\boldsymbol{cols}} \in \mathbb{R}^{B \times d_h}$
            \STATE Load $\boldsymbol{V}_{\text{chip}}
        \gets \boldsymbol{V}^{\boldsymbol{cols}} \in \mathbb{R}^{B \times d_h}$
            \STATE $\boldsymbol{S} \gets \tau\boldsymbol{Q}_{\text{chip}}\boldsymbol{K}_{\text{chip}}^T$
            \STATE $\boldsymbol{S} \gets \mathrm{mask}(\boldsymbol{S})$
            \STATE $\boldsymbol{m}^i_{new} \gets \mathrm{max}(\boldsymbol{m}^i, \mathrm{rowmax}(\boldsymbol{S})) \in \mathbb{R}^{B}$
            \STATE $\boldsymbol{S} \gets \boldsymbol{S} - \boldsymbol{m}^i_{new}$
            \STATE $\boldsymbol{P} \gets \mathrm{exp}(\boldsymbol{S})$
            \STATE $\boldsymbol{l}^i_{new} \gets \mathrm{rowsum}(\boldsymbol{S}))$
            \STATE $\boldsymbol{\alpha} \gets \mathrm{exp}(\boldsymbol{m}^i - \boldsymbol{m}^i_{new})$
            \STATE $\boldsymbol{l}^i \gets \boldsymbol{\alpha}\boldsymbol{l}^i + \boldsymbol{l}^i_{new}$
            \STATE $\boldsymbol{O}_{\text{chip}} \gets \boldsymbol{\alpha}\boldsymbol{O}_{\text{chip}} + \boldsymbol{P}\boldsymbol{V}_{\text{chip}}$
            \STATE $j \gets j + B$
        \ENDWHILE

        \LineComment{Write outputs}
        \STATE $\boldsymbol{O}_{\text{chip}} \gets \mathrm{diag}(\boldsymbol{l}^i)^{-1}\boldsymbol{O}_{\text{chip}}$
        \STATE Save $\boldsymbol{O}_i \gets \boldsymbol{O}_{\text{chip}}$
    \ENDFOR

\end{algorithmic}
\end{algorithm}
\end{minipage}
\end{figure}

\begin{table}[ht]
    \centering
    \small
    \caption{Comparison of generated results by different methods using LLaMA-3-8B-Instruct-262K in summarization task~\cite{zhang2024inftybench}.}
    \begin{tabular}{p{2cm}|p{10cm}}
    \toprule
         Methods & Summary  \\
        \midrule
        Original &  The story follows the Bronwyn family and their guests as they prepare for a day trip to the Kindergarten, a mysterious destination that holds great significance for them. The story is told from multiple perspectives, including that of Pete Lucy, who is staying with the family and is struggling to find her place in their lives. As the day progresses, the family members prepare for their trip, and Pete reflects on her own past experiences with the Bronwyns. The story ends with the family setting off on their journey, leaving Pete alone on the lawn to contemplate her own feelings and the meaning of life. The Kindergarten is a symbol of the family's past and their memories, and the story explores themes of nostalgia, loss, and the passage of time. Ultimately, the story suggests that even though the past cannot be changed, it can still be celebrated and remembered.\\ \midrule
        StreamingLLM & The story follows a family on a boat trip to the Kindergarten, a school for boys. The father, Mr Bronwyn, is reading a book and seems to be in a trance. The children, Seb and Annalisa, are excited about the trip but also a little scared. They pass by a group of fishermen and Mr Bronwyn shares out sandwiches. They come across a spot where three men were drowned and Mr Bronwyn takes out his watch, making a calculation. They reach the Kindergarten and Mr Bronwyn jumps out of the boat, holding a parcel. The story ends with Pete Lucy, who has been watching the family from the shore, feeling relieved that the trip is over and that Mr Bronwyn has reached his destination. The story is told from multiple perspectives, including Pete Lucy's, who is watching the family from the shore and reflecting on his own life and relationships. \\ \midrule
        StreamingLLM w/ dilated & A ( ( ( ( ( ( ( ( ( ( ( ( ( ( ( ( ( ( ( ( ( ( ( ( ( ( ( ( ( ( ( ( ( ( ( ( ( ( ( ( ( ( ( ( ( ( ( ( ( ( ( ( ( ( ( ( ( ( ( ( ( ( ( ( ( ( ( ( ( ( ( ( ( ( ( ( ( ( ( ( ( ( ( ( ( ( ( ( ( ( ( ( ( ( ( ( ( ( ( ( ( ( ( ( ( ( ( ( ( ( ( ( ( ( ( ( ( ( ( ( ( ( ( ( ( ( ( ( ( ( ( ( ( ( ( ( ( ( ( ( ( ( ( ( ( ( ( ( ( ( ( ( ( ( ( ( ( ( ( ( ( ( ( ( ( ( ( ( ( ( ( ( ( ( ( ( ( ( ( ( ( ( ( ( ( ( ( ( ( ( ( ( ( ( ( ( ( ( ( ( ( ( ( ( ( ( ( ( ( ( ( ( ( ( ( ( ( ( ( ( ( ( ( ( ( ( ( ( ( ( ( ( ( ( ( ( ( ( ( ( ( ( ( ( ( ( ( ( ( ( ( ( ( ( ( ( ( ( ( ( \\ \midrule
        StreamingLLM w/ strided & The GraphicsUnit36.OOAKA new ( ( ( ( ( ( ( ( ( ( ( ( ( ( ( ( ( ( ( ( ( ( ( ( ( ( ( ( ( ( ( ( ( ( ( ( ( ( ( ( ( ( ( ( ( ( ( ( ( ( ( ( ( ( ( ( ( ( ( ( ( ( ( ( ( ( ( ( ( ( ( ( ( ( ( ( ( ( ( ( ( ( ( ( ( ( ( ( ( ( ( ( ( ( ( ( ( ( ( ( ( ( ( ( ( ( ( ( ( ( ( ( ( ( ( ( ( ( ( ( ( ( ( ( ( ( ( ( ( ( ( ( ( ( ( ( ( ( ( ( ( ( ( ( ( ( ( ( ( ( ( ( ( ( ( ( ( ( ( ( ( ( ( ( ( ( ( ( ( ( ( ( \\ \midrule
        Ours w/ static & def solverome2 def solverome2 def solverome2 def solverome2 def solverome2 def solverome2 def solverome2 def solverome2 def solverome2 def solverome2 def solverome2 def solverome2 def \\ \midrule
        Ours & The story follows the Bronwyn family and their guests as they prepare for a trip to the Kindergarten, a remote island off the coast of Scotland. The family is made up of Mr. Bronwyn, a philosopher, and his wife, Mrs. Bronwyn, who is known for her beauty and her ability to surround and protect her family. The guests include Kevin Emma, a young man who is studying for a fellowship, and Castor Kiki, a botanist who is also a widower. The story explores themes of love, marriage, and the search for meaning in life. It also touches on the idea of the other, or the person who is different from oneself and cannot be understood or judged by conventional standards. Ultimately, the story is about the search for truth and the desire to connect with others on a deeper level. \\
        \bottomrule
    \end{tabular}
    \label{tab:case_summary}
\end{table}

\begin{table}[ht]
    \centering
    \small
    \caption{Comparison of generated results by different methods using LLaMA-3-8B-Instruct-262K in KV Retrieval task~\cite{zhang2024inftybench}.}
    \begin{tabular}{p{2cm}|p{10cm}}
    \toprule
         Methods & Summary  \\
        \midrule
        Original & 1. Ground Truth: 5e6b7b90-710d-4953-9b18-3e96b2cadbf2 \newline
        Prediction: "5e6b7b90-710d-4953-9b18-3e96b2cadbf2" \newline
        
        2. Ground Truth: f1ec3dd8-b6e7-4af5-a589-fda0b7693f93 \newline
        Prediction: "f1ec3dd8-b6e7-4af5-a589-fda0b7693f93"  (a string of length 32). \\ \midrule
        StreamingLLM & 1. Ground Truth: 5e6b7b90-710d-4953-9b18-3e96b2cadbf2 \newline
        Prediction: 2b114db0-d87e-42d2-9b4c-0b1f115976ad.assistant \newline      
        
        2. Ground Truth: bbf3fa73-6217-4e6e-923c-8349bd286c3d \newline
        Prediction: "d829ce18-6339-4f77-8c04-31fc7ec33619". \\ \midrule
        StreamingLLM w/ dilated & 1. Ground Truth: 5e6b7b90-710d-4953-9b18-3e96b2cadbf2 \newline
        Prediction: 5 ( ( ( ( ( ( ( ( ( ( ( ( ( ( ( ( ( ( ( ( ( ( ( ( ( ( ( ( ( ( ( ( ( ( ( ( ( ( ( ( ( ( ( ( ( ( ( ( ( ( \newline
        
        2. Ground Truth: f1ec3dd8-b6e7-4af5-a589-fda0b7693f93 \newline
        Prediction: "def solverome2 def solverome2 def solverome2 def solverome2 def solverome2 def solverome2 def solverome2 def solverome2 def solverome2 def solverome2  \\   \midrule
        StreamingLLM w/ strided & 1. Ground Truth: 5e6b7b90-710d-4953-9b18-3e96b2cadbf2 \newline
        Prediction: "def solverome2 def solverome2 def solverome2 def solverome2 def solverome2 \newline
        
        2. Ground Truth: f1ec3dd8-b6e7-4af5-a589-fda0b7693f93 \newline
        Prediction: "0 ( ( ( ( ( ( ( ( ( ( ( ( ( ( ( ( ( ( ( ( ( ( ( ( ( ( ( ( (  \\ \midrule
        Ours w/ static & 1. Ground Truth: 5e6b7b90-710d-4953-9b18-3e96b2cadbf2 \newline
        Prediction: "def solverome2 def solverome2 def solverome2 def solverome2 def \newline
        
        2. Ground Truth: f1ec3dd8-b6e7-4af5-a589-fda0b7693f93 \newline
        Prediction: "def solverome2 def solverome2 def solverome2 def solverome2 def
        \\ \midrule
        Ours &  1. Ground Truth: 5e6b7b90-710d-4953-9b18-3e96b2cadbf2 \newline
        Prediction: "5e6b7b90-710d-4953-9b18-3e96b2cadbf2" \newline
        
        2. Ground Truth: f1ec3dd8-b6e7-4af5-a589-fda0b7693f93 \newline
        Prediction: "f1ec3dd8-b6e7-4af5-a589-fda0b7693f93"  (a string of length 32).
        \\  
        \bottomrule
    \end{tabular}
    \label{tab:case_kv}
\end{table}

\end{document}